\documentclass[runningheads]{llncs}
\usepackage{graphicx}
\usepackage{tikz}
\usepackage{comment}
\usepackage{amsmath,amssymb,bm} % define this before the line numbering.
\usepackage{color}
\usepackage[inline]{enumitem}
\usepackage{enumerate}
\usepackage[super]{nth}

\definecolor{citecolor}{RGB}{34,139,34}

\usepackage{epsfig}
\usepackage{bbm}
\usepackage{xcolor}
\usepackage{mathtools}

\usepackage{tabularx}
\usepackage{multirow}
\usepackage[pagebackref=true,breaklinks=true,colorlinks,citecolor=citecolor,bookmarks=false]{hyperref}

\usepackage{booktabs}
\usepackage{makecell}
\usepackage{pifont}

\usepackage{adjustbox}
\usepackage[normalem]{ulem}
\useunder{\uline}{\ul}{}
\usepackage{rotating}

\usepackage{listings}
\usepackage{amsfonts}

\usepackage{fontenc}
\usepackage{multicol}
\usepackage{array}

\newcolumntype{H}{>{\setbox0=\hbox\bgroup}c<{\egroup}@{}}

\newcommand{\detr}{\mbox{DETR}}
\newcommand{\vistr}{\mbox{VisTR}}
\newcommand{\evis}{\mbox{DeVIS}}

\newcommand{\ie}{i.e.}
\newcommand{\eg}{e.g.}

\usepackage{pifont}% http://ctan.org/pkg/pifont

\renewcommand{\Sigma}{\mathfrak{S}}

\def\eqref#1{equation~\ref{#1}}

\def\1{\bm{1}}

\def\vp{{\bm{p}}}
\def\vp{{\bm{p}}}

\def\vx{{\bm{x}}}

\def\vz{{\bm{z}}}

\def\mW{{\bm{W}}}

\DeclareMathAlphabet{\mathsfit}{\encodingdefault}{\sfdefault}{m}{sl}
\SetMathAlphabet{\mathsfit}{bold}{\encodingdefault}{\sfdefault}{bx}{n}

\def\sR{{\mathbb{R}}}

\def\emA{{A}}

\begin{document}
% \linenumbers
\pagestyle{headings}
\mainmatter
\def\ECCVSubNumber{523}  % Insert your submission number here

\newif\ifarxiv
\arxivtrue

\title{DeVIS: Making Deformable Transformers \\ Work for Video Instance Segmentation} 

\begin{comment}
\titlerunning{ECCV-22 submission ID \ECCVSubNumber} 
\authorrunning{ECCV-22 submission ID \ECCVSubNumber} 
\author{Anonymous ECCV submission}
\institute{Paper ID \ECCVSubNumber}
\end{comment}
%******************

% \begin{comment}
\titlerunning{DeVIS}
\author{Adrià Caelles\inst{1} \and Tim Meinhardt\inst{2} \and
Guillem Brasó\inst{2} \and Laura Leal-Taixé\inst{2}}
\authorrunning{Caelles et al.}
\institute{Technical University of Catalonia \\
\email{adria.caelles@estudiantat.upc.edu}
\and
Technical University of Munich\\
\email{\{tim.meinhardt,guillem.braso,leal.taixe\}@tum.de}}
% \end{comment}

%******************

\maketitle

\begin{abstract}
% Near-online Video Instance Segmentation effectively reason about clip information while being able to run on arbitrary clip lengths.

% We present SOTA results (we don't have them yet though)  while running near-online manner, making clear our superiority among offline algorithms. 

% Our method is specially conceived to work in this domain leveraging a novel Temporal Multi-Scale deformable attention.

Video Instance Segmentation (VIS) jointly tackles multi-object detection, tracking, and segmentation in video sequences.
In the past, VIS methods mirrored the fragmentation of these subtasks in their architectural design, hence missing out on a joint solution.
Transformers recently allowed to cast the entire VIS task as a single set-prediction problem.
Nevertheless, the quadratic complexity of existing Transformer-based methods requires long training times, high memory requirements, and processing of low-single-scale feature maps.
Deformable attention provides a more efficient alternative but its application to the temporal domain or the segmentation task have not yet been explored.

In this work, we present Deformable VIS (\evis{}), a VIS method which capitalizes on the efficiency and performance of deformable Transformers.
To reason about all VIS subtasks jointly over multiple frames, we present temporal multi-scale deformable attention with instance-aware object queries.
We further introduce a new image and video instance mask head with multi-scale features, and perform near-online video processing with multi-cue clip tracking.
%
% \evis{} benefits from comparatively small memory as well as training time requirements, and achieves state-of-the-art results on the YouTube-VIS 2019 and 2021, as well as the challenging OVIS dataset.
\evis{} reduces memory as well as training time requirements, and achieves state-of-the-art results on the YouTube-VIS 2021, as well as the challenging OVIS dataset.

Code is available at~\url{https://github.com/acaelles97/DeVIS}.


\end{abstract}

%************* INTRODUCTION NARRATIVE & MAIN IDEAS *************:
% What is VIS -> How are VIS algorithms classified -> What characterizes each type and why we like the near-online approach -> DeTr first to apply Transformer on Vision tasks->  What does DeTr suffer from and how Deformable DeTR solves it -> VisTR first to expand DeTR (transformers) to video -> Similarly to DeTR, what does VisTR suffers from -> How other VIS algorithms have tried to solve it (IFC & SeqFormer) -> Our solution to the problem and why we think is better :)
%***************************************************************

%---------------------------------------------------------------
% Why I think it is important to highlight that our model has been conceived for the near-online domain?
% The spatial limitation of the temporal deformable attention on the transformer encoder can then be seen as a part of the architecture, not a limitation.
% I think it is also important to put in evidence why we can not take our model and bring it to the offline domain with larger clip sizes
%---------------------------------------------------------------

\section{Introduction}
Video Instance Segmentation (VIS) simultaneously aims to detect, segment, and track multiple classes and object instances in a given video sequence. %
Thereby, VIS provide rich information for scene understanding in applications such as autonomous driving, robotics, and augmented reality.
% In comparison to single image object recognition, the temporal domain plays a beneficial role for the first three subtasks.
%
%In particular, tracking and segmentation during occlusions makes VIS a challenging video recognition task which offers rich information for scene understanding in applications such as autonomous driving, robotics, and augmented reality.

% The VIS task was only recently introduced~\cite{Yang2019vis} to the computer vision community and 
%
Early methods~\cite{Yang2019vis,sip_mask,cross_vis} took inspiration from the multi-object tracking (MOT) field by applying tracking-by-detection methods to VIS, \eg, ~\cite{Yang2019vis} extends Mask R-CNN~\cite{he2017mask} with an additional tracking head.
The frame-by-frame mask prediction and track association allow for real-time processing but fail to capitalize on temporal consistencies in video data.
Hence, more recent VIS methods~\cite{stem_seg,mask_prop,prop_reduce,stmask,sg_net} moved towards an offline or near-online processing of clips by treating instance segmentations as 3D spatio-temporal volumes.

The recent success of Transformers~\cite{attention_is_all_you_need} in object recognition
%
%initiated by the \detr{}~\cite{DETR} object detector
%
inspired a new generation of VIS approaches.
%
%~\cite{vistr,IFC} \gui{there are only 2 published right?}. %has also left its mark on the VIS community, and.
%However, one way or the other all existing methods are limited by the quadratic complexity of vanilla attention with respect to its input \gui{size}.
%
Both \vistr{}~\cite{vistr} and IFC~\cite{IFC} deploy an encoder-decoder Transformer architecture, and closely mirror \detr{}'s~\cite{DETR} approach to object detection by formulating the VIS task as a set-prediction problem.
%The~\cite{vistr,IFC} methods \gui{I'd drop 'The' ... and 'methods'} both deploy an encoder-decoder Transformer architecture by formulating the VIS task as a set-prediction problem. \gui{maybe add:, closely mirroring DETR's approach to object detection.}
%
In this paradigm, instance masks are obtained in a single end-to-end trainable forward pass for all frames in a clip.
%
%avoiding any post-processing heuristics.
%
While their formulation is simple and appealing, both methods are limited by the quadratic complexity of full attention. 
%\lau{I would not say this: "and hence inherit all limitations of~\cite{DETR}." I mean DETR has other limitations, but indeed it was not created for VIS, so I would rather point to the quadratic complexity of full attention, which for videos it just explodes as you say below.}
%
%with respect to its input size.
%
\vistr{} achieves communication between frames by concatenating and encoding the pixels of all frames jointly. 
Such a multi-frame processing only amplifies the expensive attention computation and makes~\cite{vistr} suffer from long training times and high memory requirements.
%
% These limitations, which were already present in~\cite{DETR}, are only aggravated through the added burden of processing multiple frames at once.
%Similar to~\cite{DETR} but even more emphasized through the processing of multiple frames at once, ~\cite{vistr} suffers from very long training times, if not applied to low input resolutions.
%
IFC~\cite{IFC} tries to mitigate these issues by introducing inter-frame memory tokens to encode each frame individually.
However,~\cite{IFC} still relies on full attention for single frames and hence inherits the limitations of~\cite{DETR,vistr} to only process low- and single-scale feature maps.

\begin{figure}[t]
    \centering
    \includegraphics[width=1\textwidth]{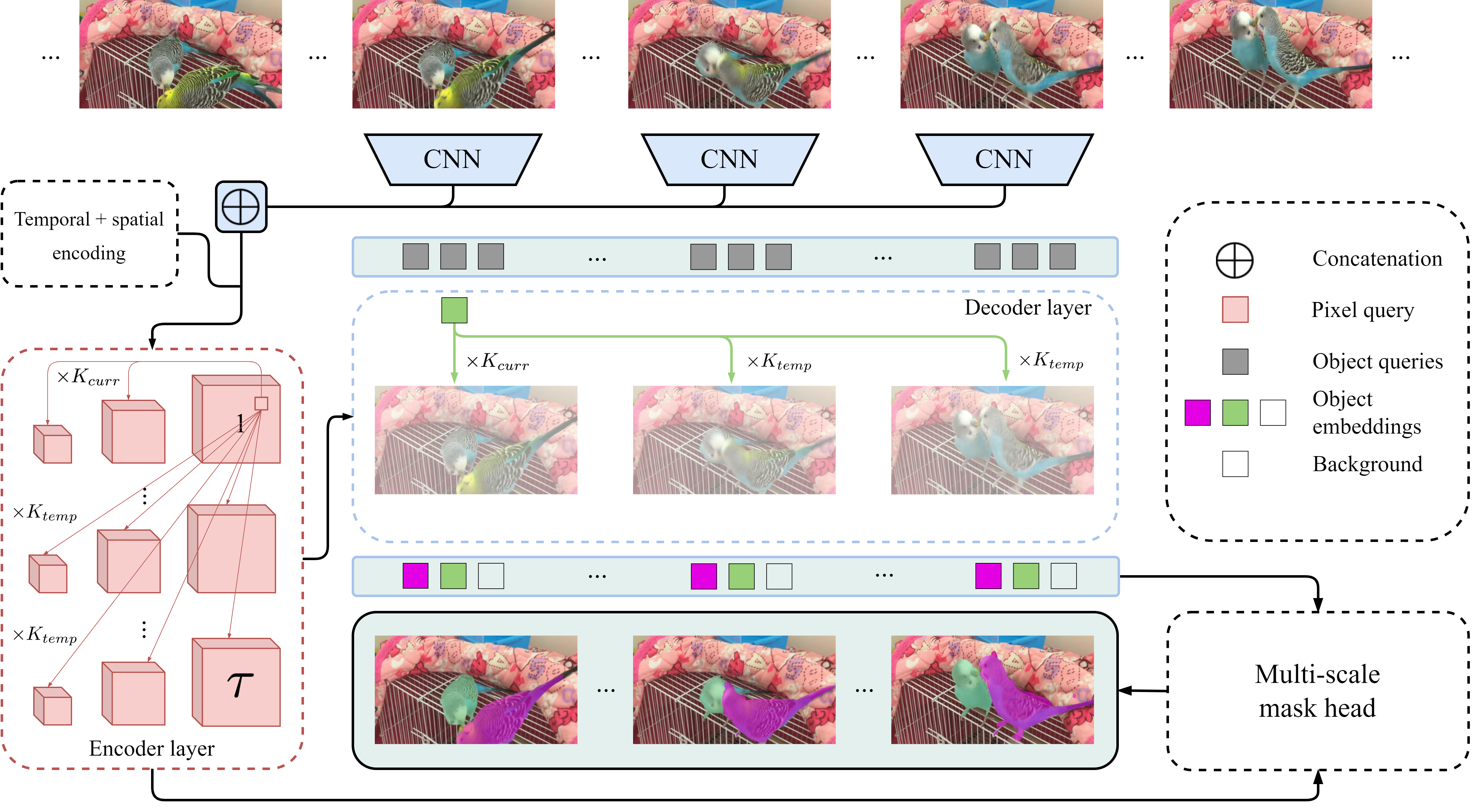}
      %  \vspace{-0.4cm}
    \caption{
    An overview of our~\textbf{~\evis{} method} which applies temporal multi-scale deformable attention in a Transformer encoder-decoder architecture.
    The encoder computes deformable attention between pixels across scales and frames in a given clip without suffering from quadratic complexity of full attention.
    In the decoder, object queries attend to multiple frames, thereby providing consistent identity predictions.
    }
    \label{fig:method}
   % \vspace{-0.4cm}
\end{figure}

Deformable attention~\cite{deformable_detr} resolves many of \detr{}`s efficiency and performance issues, and circumvents the quadratic computational complexity for its inputs by subsampling attention keys around a spatial reference point assigned to each query.
%
%Deformable attention~\cite{deformable_detr} resolved many of \detr{}`s efficiency and performance issues for object detection by relaxing the quadratic complexity and only sampling a subset of attention keys around the spatial reference point of each query.
%
As shown in our experiments, a naive~\textit{deformabilization} of~\vistr{},~\ie, replacing full attention with deformable attention in the encoder-decoder of Figure~\ref{fig:method}, does not achieve satisfactory train time nor segmentation performance.
This is largely due to two problems: (i) learnable reference point offsets and attention weights introduce an unfeasible amount of new parameters for long clip sizes, and (ii) attention with spatially local reference points is not well-suited for the detection and tracking of objects moving through a sequence.
%
% \begin{enumerate*}[label=(\arabic*)]
%     \item learnable reference point offsets and attention weights introduce an unfeasible amount of new parameters for long clip sizes, and 
%     \item attention with spatially local reference points is not well-suited for the detection and tracking of objects moving through space and time in a sequence.
% \end{enumerate*}

% To make deformable attention work for VIS and mitigate the aforementioned training time and single-low-scale feature map issues, we present~\textit{Deformable VIS} (\evis{}), a Transformer encoder-decoder approach which applies temporal deformable attention over multiple frames.
To make deformable attention work for VIS, we present~\textit{Deformable VIS} (\evis{}), a Transformer encoder-decoder which applies temporal deformable attention over multiple frames.
The reduction in computational complexity reduces training time and makes the processing of high-res feature maps at multiple scales feasible.
Furthermore, we motivate the alignment of reference points for decoder object queries individually for each object instance.
Our newly proposed image and video instance segmentation head takes full advantage of multi-scale features and improves mask quality significantly.
%Furthermore, we present a new instance segmentation mask head which boosts segmentation performance by taking full advantage of the encoded multi-scale features maps. \gui{Also, do we wanna mention that this contribution is also for general instance segmentation in the end?}
%
To run sequences with arbitrary lengths, we also introduce an improved multi-cue clip tracking.
The presented~\evis{} method achieves state-of-the-art results on the challenging YouTube-VIS~\cite{Yang2019vis} 2021 and OVIS~\cite{ovis} datasets and substantially reduces training time with respect to~\vistr{}.

% \tim{clarify that the original VIS paper already used multiple cues for cost-based tracking. we are only the first ones applying something like this to a encoder-decoder Transformer architecture.}
% \adria{I think that we need to clarify the following: We are the first to apply multi-cue to cost based matching on overlapping frames during inference. This is different from online-tracking scenario in which multi-cue was introduced long ago, on MaskTrack R-CNN. Note that on MaskTrack R-CNN, the highest multi-cue score is picked for each previous instance, there is no hungarian matching going on there.}

In summary, our key \textbf{contributions} are:

\begin{itemize}
\item We present Deformable VIS (\evis{}), a VIS method which introduces temporal multi-scale deformable attention and instance-aware object queries. %  reference point sampling for

\item We present a new instance mask prediction head which takes full advantage of deformable attention and encoded multi-scale feature maps.

\item Our improved multi-cue clip tracking incorporates mask and class information to connect overlapping clips to sequences of arbitrary length.

\item Our method provides efficient training and achieves state-of-the-art performance on YouTube-VIS 2021 and the challenging OVIS dataset. %while requiring less training time and memory than previous Transformer-based VIS methods.

\end{itemize}

\section{Related work}

We discuss VIS methods following the progression from tracking-by-detection to clip-level approaches culminating in modern Transformer-based architectures.

\noindent \textbf{Tracking-by-detection.}
The inception of the VIS task with the YouTube-VIS 2019~\cite{Yang2019vis} dataset also created Mask-Track R-CNN~\cite{Yang2019vis}.
As it is common in the multi-object-tracking community, Mask-Track R-CNN processes sequences frame-by-frame in an online tracking-by-detection manner.
To this end,~\cite{Yang2019vis} extends Mask R-CNN~\cite{he2017mask} with a tracking branch which allows it to not only detect and segment objects, but also assign instance identities via similarity matching of instance embeddings in the current frame and a memory queue.
SipMask~\cite{sip_mask} applies the same tracking head but with a single-stage detector and light-weight spatial mask preservation module.
The crossover learning scheme of CrossVIS~\cite{cross_vis} allows for a localization of pixel instance features in other frames.

As a clip-level method,~\evis{} processes sequences in clips, which greatly improves quality and robustness of instance mask as well as identity predictions.

\noindent \textbf{Clip-level.}
Clip-level processing allows for offline or near-online VIS methods.
The latter clip a sequence into multiple parts and hence rely on an additional clip tracking step.
The STEm-Seg~\cite{stem_seg} method took inspiration from offline trackers and is the first to model object instances as 3D spatio-temporal volumes by predicting pixel embeddings with Gaussian variances.
For a hybrid tracking-by-detection and clip-level method, the authors of MaskProp~\cite{mask_prop} extend Mask R-CNN with a mask propagation branch that operates between all frames in a video clip.
The~\textit{Produce-Reduce} heuristic applied in SeqMask-RCNN~\cite{prop_reduce} generates object instance proposals and reduces redundant identities based on multiple key frames.
STMask~\cite{stmask} and SG-Net~\cite{sg_net}, on the other hand, move beyond Mask R-CNN by applying improved one-stage detection methods.

Albeit their early success, these clip-level methods usually tackle the detect, segment, and track subtasks with separately trainable multi-stage pipelines.
Our~\evis{} approach enjoys the advantages of clip-level methods but in a unified and end-to-end trainable manner through the application of Transformers.

\noindent \textbf{Clip-level with Transformers.}
The~\vistr{}~\cite{vistr} method introduced Transformers~\cite{attention_is_all_you_need} to VIS by extending the \detr{}~\cite{DETR} object detector to the temporal domain.
Its unified Transformer encoder-decoder architecture concatenates and encodes all frames in a clip by computing attention between all pixels.
The decoder reasons about detection and tracking via multi-frame cross-attention between object queries and the encoded pixels, and produces instance masks with a subsequent segmentation head.
%
% The multi-frame attention allows to reason about detection, tracking, and segmentation in a unified way.
%
To avoid the expensive computation multi-frame pixel attention, the authors of~\cite{IFC} encode each frame separately and introduce memory tokens for a high-level inter-frame communication.
%
% These allow for communication between frames on a higher level by restricting pixel encoding to each frame.

Our proposed~\evis{} method mitigates the aforementioned efficiency issues while still benefiting from the simultaneous encoding of multiple frames at once through the application of temporal multi-scale deformable attention with instance-aware reference point sampling.
We further propose a new segmentation head which takes full advantage of the Transformer encoded multi-scale feature maps, and an improved multi-cue clip-tracking.

\section{\evis{}}
In this section, we present \textit{Deformable VIS} (\evis{}), a near-online end-to-end trainable Transformer encoder-decoder architecture for VIS.
We details its key components: (i) Temporal multi-scale deformable attention on a (ii) Transformer architecture with instance-aware object queries, (iii) a new multi-scale deformable mask head, and (iv) multi-cue clip tracking.

\subsection{Clip-level VIS with Transformers}
\label{sec:clip_level_vis_trans}

% Extending the concept of~\cite{DETR} to the temporal domain, the authors of~\vistr{}~\cite{vistr} presented the first application of a Transformer encoder-decoder architecture for the VIS task.
%
% To this end, they applied the concept of the \detr{}~\cite{DETR} object detector to the temporal domain by simultaneously processing a clip of frames.
%
In this section, we present an overview of our~\evis{} method, shown in Figure~\ref{fig:method}, which follows the general Transformer-based clip processing pipeline of~\cite{vistr}.

\noindent \textbf{Clip-level VIS.} Given a video with $T$ frames, the goal of VIS is to detect, track, and segment all $K$ objects in a sequence.
This is achieved by providing a set of instance predictions $\Upsilon = \{y_{i,t}\}$ with $y_{i,t} = \{c_{i,t}, b_{i,t}, m_{i,t}\}$.
A single prediction consists of the class $c$, bounding box $b$, and mask $m$ for instance identity $i$ at frame $t$.
The VIS task expects a constant $c_{i,t}$ for all $t$.
If an object instance with identity $j$ is not present for the entire sequence, the final set $\Upsilon$ can include less than $T$ predictions with $y_{j,t}$.
A clip-level VIS method processes clips with $\tau$ frames and first provides subsets of instance predictions $\Upsilon_k$ with distinct sets of identities.
Usually, clips include overlapping frames to perform a final clip tracking/stitching step, which is responsible for merging identities of overlapping clip instance predictions.

\noindent \textbf{VIS with Transformers.}
To generate a set of clip predictions $\Upsilon_k$, Transformer encoder-decoder methods first extract feature maps for each frame independently with a convolutional neural network (CNN).
Feature maps are then concatenated to form a clip and temporal-positional encoding is added.
Treating each pixel as an input query, the subsequent Transformer encoder shares information spatially and temporally between frames via self-attention~\cite{attention_is_all_you_need}.
A set of learned object queries~\cite{DETR} computes self-attention and cross-attention with all encoded pixels in the Transformer decoder.
The total amount of object queries is equally distributed over the frames, hence, a query attends to all pixels in the clip but is responsible for the predictions on a fixed frame.
The decoder outputs a set of object query embeddings which are passed through separate multi-layer perceptrons to predict $c_{i,t}$ and $b_{i,t}$ for each frame in the clip.
The instance mask predictions $m_{i,t}$ are obtained by computing attention maps for each object embedding, and feeding these together with the backbone feature maps into an additional instance mask head.
The entire model is trained end-to-end by matching the predicted outputs $y_{i,t}$ via a Hungarian cost matrix to the ground truth.
We refer to~\cite{DETR} and ~\cite{vistr} for more details on the matching and loss computation.
Instance identities $i$ are predicted by matching a fixed set of queries each from a different frame to the same identity during training.
At inference, all objects detected and segmented by one of these query sets are then assumed to belong to the same identity.
Intuitively, object queries in~\detr{} learn to detect objects in certain regions of the image.
For VIS, each set of queries is responsible for certain types of spatio-temporal object trajectories through the clip.

% \subsection{Temporal multi-scale deformable attention}
% \subsection{From attention to temporal multi-scale deformable attention}
% \subsection{From attention  to temporal deformable attention}
\subsection{Roadmap to temporal deformable attention}
\label{sec:temp_ms_def_att}

The authors of~\cite{deformable_detr} introduced deformable attention to~\detr{}, thereby reducing the computational footprint and training time substantially.
This allowed~\cite{deformable_detr} to improve single-image object detection performance by running the Transformer encoder-decoder on multiple feature scales.
While they only operated on single images, we present deformable attention for the \textit{temporal domain} which simultaneously captures spatial and temporal dependencies across multiple scales.

\subsubsection{Multi-Head Attention.} The original Transformer~\cite{attention_is_all_you_need} applies full attention between two sets of input queries $\Omega_q$ and keys $\Omega_k$.
An element of each of these sets is denoted by $k$ and $q$ with feature representations $\vz_q \in \sR^{C}$ and $\vx_k \in \sR^{C}$ of hidden size $C$, respectively.
We denote the set of all $\vx_k$ as $\mathbf{X}$.
The Multi-Head Attention (MHA) for query $q$ and $M$ attention heads is then computed via:

\begin{equation}
\text{MHA}(\vz_q, \mathbf{X}) = \sum_{m=1}^{M} \mW_m \big[\sum_{k\in\Omega_k} \emA_{mqk} \cdot \mW'_m \vx_k \big],
\label{eq:mha}
\end{equation}

with learnable weight matrices $\mW'_m \in \sR^{C_v \times C}$ and $\mW_m \in \sR^{C \times C_v}$ where $C_v = C/M$.
The attention weights $\emA_{mqk}$ are computed via dot product between $\vz_q$ and $\vx_k$ and are normalized over all keys $\sum_{k\in\Omega_k} \emA_{mqk} = 1$.
The case where $\Omega_q = \Omega_k$ is usually referred to as self-attention.
The Transformer encoder in~\detr{}~\cite{DETR} applies self-attention, where every entry in the feature map corresponds to a query.
Due to the computation of $\emA_{mqk}$, which scales quadratically with the number of queries/keys, it is only feasible to run~\cite{DETR} on a single feature scale.

\subsubsection{Deformable attention.}
To mitigate~\detr{}'s computational issues around attention, the authors of~\cite{deformable_detr} suggest deformable attention which works on subsets of queries and weight computation of linear complexity.
To this end, each query $q$ is assigned a reference point $\vp_q \in \sR^2$ in the feature map domain.
A subset of $K$ queries is sampled around the reference point based on sample offsets $\Delta\vp_{mqk}$ which are learnable via linear projection over the query feature $\vz_q$.
For simplicity we omit the summation over multiple attention heads and denote the resulting \textit{Deformable Attention} (DA) for a single attention head $m$ as:

\begin{equation}
\text{DA}(\vz_q, \vp_q, \mathbf{X}) = \mW_m \big[\sum_{k=1}^{K} \emA_{mqk} \cdot \mW'_m \vx(\vp_q + \Delta\vp_{mqk})\big].
\label{eq:dmha}
\end{equation}

The attention weights are normalized over the sample points ${\sum_{k\in\Omega_k} \emA_{mqk} = 1}$ and also obtained via linear projection which avoids the expensive computation of dot product between queries.
Furthermore,~\cite{deformable_detr} present a multi-scale version of Equation~\ref{eq:dmha} which computes deformable attention across feature maps.

\subsubsection{Temporal multi-scale deformable attention.}
To encode spatio-temporal dependencies, which are crucial for for VIS, we present multi-scale deformable attention for the temporal domain.
That is, the sets of queries and keys include pixels from multiple scales and all frames in a clip.
Hence, we re-define $\mathbf{X} = \{\mathbf{X}^{l}\}_{l=1}^{L} $ to be the stack of $L$ multi-scale backbone features with $\tau$ frames where $\mathbf{X}^{l} \in \sR^{C \times \tau \times H_l \times W_l}$.
Intuitively, a query $q$ from frame $t$ has the ability to compute attention with sampled keys from all frames and feature levels in a clip.
In comparison to full attention between all pixels, the sampling with offsets around a query`s reference point $\hat{\vp}_q$ reduces the computational effort substantially.
The number of keys is independent of the input resolution and only scales linearly with the number of feature scales and frames.
We define Temporal Multi-Scale Deformable Attention (TMSDA) module for a single $m$ over a clip as:

% \begin{align}
% \label{eq:temporal_ms_deform_attn_fun}
% \text{TMSDA}(\vz_q, \hat{\vp}_q, \{ \mathbf{X}_t \}_{t=1}^{\tau}) 
% = \sum_{m=1}^{M} \mW_m \big[\sum_{t=1}^{\tau} \sum_{l=1}^{L} \sum_{k=1}^{K(t)} \emA_{m lqk} \cdot \mW'_m \vx_{t}^{l}(\phi_{l}(\hat{\vp}_q) + \Delta\vp_{m\tau lqk})\big],
% \end{align}

\begin{equation}
\label{eq:tmsdmha}
\text{TMSDA}(\vz_q, \hat{\vp}_q, \mathbf{X}) = \mW_m \big[\sum_{t=1}^{\tau} \sum_{l=1}^{L} \sum_{k=1}^{K(t)} \emA_{mtlqk} \cdot \mW'_m \vx^{lt} (\phi_{l}(\hat{\vp}_q) + \Delta\vp_{mt lqk})\big].
\end{equation}

As in~\cite{deformable_detr}, each reference point is represented with normalized coordinates $\hat{\vp}_q \in [0, 1]^2$ and re-scaled by $\phi_{l}$ to allow for a sampling across feature maps $l$ with different resolutions.
%
% Furthermore, the function $\phi_{l}(\hat{\vp}_q)$ re-scales the normalized coordinates $\hat{\vp}_q$ to the input feature map of the $l$-th level.
%
The scalar attention weight $\emA_{mtlqk}$ is normalized by ${\sum_{t=1}^{\tau} \sum_{l=1}^{L} \sum_{k=1}^{K(t)} \emA_{mtlqk} = 1}$. 
We introduce $K(t)$ which adapts the number of keys sampled from a given frame and present two scenarios depending on whether a query $\vz_q$ samples keys from its corresponding or other temporal frames:
\begin{equation}
K(t)=\begin{cases}
          K_{curr}  &\text{if}\ \vz_q \in \mathbf{Z}^{lt}  \\
          K_{temp}  &\text{else}. \\
     \end{cases}
\end{equation}
Adding the temporal dimension allows each query to simultaneously sample keys from all feature scales as well as its current, and temporal frames.
Such a design is particularly beneficial for a consistent detection and identity prediction of objects moving and changing size over the sequence.
$K_{temp} = 0$ removes all temporal connections and reverts back to the original deformable attention from~\cite{deformable_detr}.

In the following paragraphs, we give further details on how our temporal deformable attention is applied in our Transformer encoder-decoder architecture.
%
% The reference point position $\hat{\vp}_q$ used when $\tau != 0$ (temporal frames) is different for the encoder self-attention and the decoder cross-attention mechanism and we detail it the corresponding section. 
% %
% How much further each query can sample from is controlled by the temporal window size $\omega$. 
% %
% The temporal window allows to control the temporal receptive field of the deformable attention mechanism. 
% %
% We have preserved the total number of different feature levels $L$ that the keys are sampled from to be independent of $\tau$. 
% %
% We have modified though the number $K$ of keys that can be sampled to depend on $\tau$. 
% %
% This allows to balance the total number of keys sampled from the current frame versus the ones from other (past or future) frames. Note that current frame information equals temporal information when $K_0 = 2\omega K_{curr}$.
% %
% In implementation, the query feature $\vz_q$ is fed to 2 different linear projection operators:  One computes the sampling offsets and attention weights for current frame keys and its total dimensions is of $3MLK_{curr}$. 
% %
% The other one computes the sampling offsets and attention weights for keys from other frames, and its dimension is of $3M2\omega LK_{temp}$. 
%

% \todo{maybe add window formulation: The window formulation allows us to run larger clip sizes without an increase in trainable parameters and keeping computational costs low.}

\subsection{TMSDA for VIS Transformers}
\begin{figure}[t]
    \centering
    \includegraphics[width=1\textwidth]{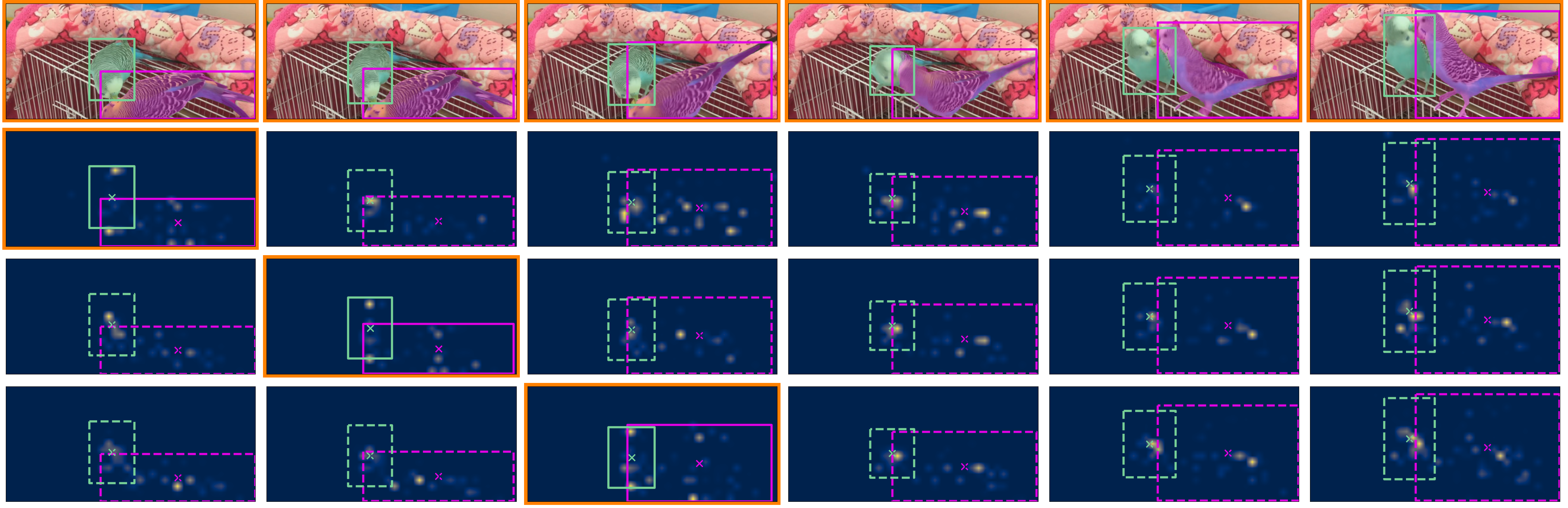}
    \caption{
    \textbf{Attention map visualization} of all frames in a clip for two object queries assigned to detect and segment object on the first, second and third frame.
    The temporal attention computed on other frames successfully follows each object and hence provides consistent identity predictions.
    Furthermore, we visualize the instance-aware reference point alignment (red dot) which adjusts the reference point of each first-frame query to the position of the respective object in the other frames.
    }
    %\vspace{-12pt}

    \label{fig:attention_qual}
\end{figure}
%\vspace{-3pt}

Following the formal introduction of Temporal Multi-Scale Deformable Attention (TMSDA), we detail its integration into our Transformer encoder-decoder and present a novel design of instance-aware reference points for object queries.

\subsubsection{Transformer encoder.}
We replace the common full self-attention between all pixels in the encoder with TMSDA as in Equation~\ref{eq:tmsdmha}. 
The sampling design of deformable attention allows each pixel to only connect to a subset of other pixels close to spatial location of the reference points $\vp_q$ for the current and other frames. 
The amount of temporal information being considered is controlled by the clip size $\tau$ and the number of temporal sampling points $K_{temp}$.
Computing deformable attention allows us to encode all $L=4$ ResNet~\cite{resnet} backbone feature scales, thereby replacing the role of a FPN~\cite{FPN}.
%
% In addition to an additive positional and feature level encoding, we follow~\cite{vistr} and apply a temporal encoding to indicate the frame position in a clip for each pixel query.
We apply an additive encoding to each pixel which indicates it spatial, temporal, and feature scale position.
%
% A major difference between our approach and~\vistr{} is the spatial locality of sampled attention keys.
%
The spatial locality of sampled keys limits the available temporal information potentially required for long clips with large object motions.
%
% Considering a pixel from frame $t$, the information around its reference point in frames far away might not be interesting due to large object motions.
%
% Hence, we do not operate on large clip sizes $\tau=36$ like~\cite{vistr} but obtain optimal results in a near-online setting, \eg, with $\tau=6$.
%
% The authors of~\cite{IFC} tackle the quadratic computational complexity in the encoder by replacing any direct temporal connections with frame-level memory tokens.
%
We believe our deformable approach is superior to~\cite{vistr,IFC} as it allows for a more fine-grained inter-frame communication on multiple scales in a unified formulation.

%
% Typical values used for this window are $4$ or $6$, which means that most of the spatial information around a given query pixel between its current frame and the temporal frames is preserved. 
% %
% This is important to note because if we wanted to sample from much further away, the information that would surround the query pixel would be much different from the one in the current frame. 
% %
% This would break the idea behind deformable attention of focusing only on most meaningful keys.

\subsubsection{Transformer decoder.}
Our Transformer decoder computes two attention steps for its object queries: self-attention and cross-attention between queries and the encoded frame features. 
%
% The with respect to the pixels comparatively small number of object queries allows for the computation of regular self-attention between the object queries.
%
For single-image detection~\cite{DETR}, the decoder self-attention helps to avoid duplicate object detections.
But for VIS Transformers, object queries also need to communicate about instance identities.
%
% As mentioned in Section~\ref{sec:clip_level_vis_trans}, the n-th query in each subset is responsible to predict the trajectory of an object over the clip.
%
As explained in Section~\ref{sec:clip_level_vis_trans}, a subset of object queries are assigned to each frame of the clip.
To extract object information with an object query from its corresponding and future as well as past frames, we compute temporal deformable multi-scale cross-attention.
In contrast to the Transformer encoder, the reference points of object queries required for the cross-attention are learnable.
%
% This allows each query to simultaneously sample multi-scale feature pixels around its reference point on its own and other frames.
%
Each object query can leverage not only meaningful, local information from its particular object on its assigned frame, but from each of the other frames of the input clip.   
%
% In addition to the self-attention between object queries,
This helps improve consistent mask and identity predictions over the sequence.
Furthermore, we apply the same bounding box refinement as~\cite{deformable_detr} at each layer of the decoder.
This allows the initial reference point, \ie, sampling area, of a query to be adapted to the currently assumed coordinates of the object bounding box.

% Similar as Deformable \detr, the sampling offset $\Delta\vp_{mlqk}$ from the decoder cross-attention mechanism is also modulated by the box size, as $({\Delta p_{mlqkx}~\hat{b}^{d-1}_{qw}}, \Delta p_{mlqky}~\hat{b}^{d-1}_{qh})$. 
% \todo{add instance query formulation as in~\cite{IFC}. this only changes the learned initialization of the queries embedding and encoding}

% \todo{I feel like this section is missing some crucial details.}

% \todo{repeat matching with GT here. and how this enforces the identities to be consistent.}

% \todo{Maybe add: Sampling modulation}

% \todo{add connection for encoder to the feature maps. since they are the queries.}

\subsubsection{Instance-aware object queries.}
%
% The assumption of consistent assignments between the aforementioned of a set of object queries to an object instance over multiple frames allows us to further adjust the sampling reference points of each query.
%
Each object query of the decoder is by design able to learn different sampling reference points for different frames in the clip.
This not only allows to distinguish between a query's assigned frame and other frames, but to adjust the reference points on other frames according to its instance identity.
Hence, we introduce~\textit{Instance-aware object queries} which exploit the identity consistency across queries to adapt their reference points on other frames to the predicted bounding boxes belonging to their respective object identity.
This instance-aware reference point sampling is applied before each Transformer decoder layer, see Figure~\ref{fig:method}, and works in conjunction with the bounding box refinement of the reference point on the query frame.
If an object query successfully predicts the object bounding box on its own frame $t$, other object queries belonging to the same instance but from other frames will benefit from an improved sampling on frame $t$. 
Instance-aware object queries provide an additional communication between object queries of different frames and improve track consistency.

\subsection{Multi-scale deformable mask head}\label{sec:def_segme_head}

\begin{figure*}
    \centering
    % \vspace{-0.5cm}
    
    \includegraphics[width=1.0\columnwidth]{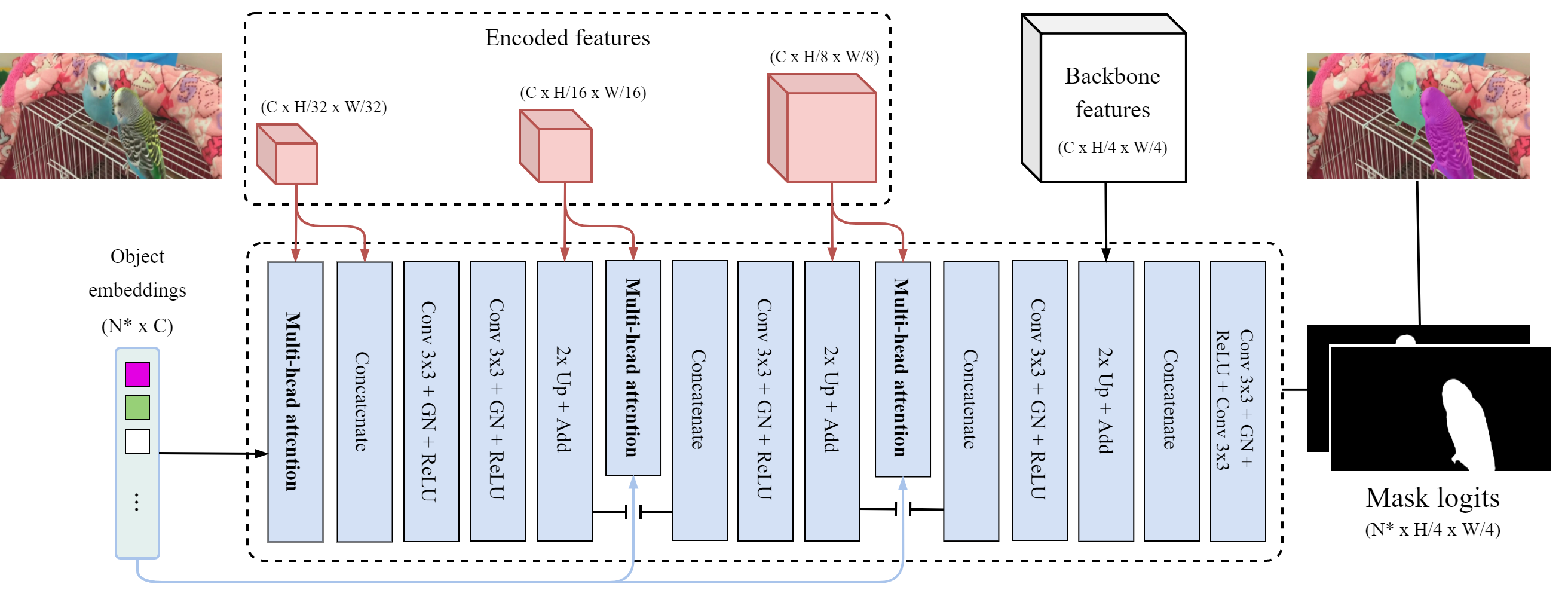}
    
    \caption{
    Overview of our new \textbf{multi-scale mask head} for video and image instance segmentation.
    The upsampling of Transformer-encoded feature maps and multi-scale attention maps boosts performance significantly.
    Attention maps are generated by computing multi-head attention between feature maps and object queries.
    We indicate the hidden size and reduced set of object queries with N$^{\ast}$ and C, respectively.
    New connections to the decoder (blue) and encoder (red) are shown colored.
     }
    
    \label{fig:arch_mask-head}
\end{figure*}

Processing multi-scale features currently only benefits the detection and tracking performance of the Transformer encoder-decoder.
We further explore the potential of Transformer encoded feature maps for mask prediction, and present a new multi-scale deformable instance mask segmentation head applicable to both image and video segmentation.

% The authors of Deformable \detr{}\cite{deformable_detr} refrained from exploring the potential of deformable attention for the mask head.

% Note that our head is not only applicable to VIS, but it can also be applied for single-image instance segmentation with very competitive results when used in conjunction with \cite{deformable_detr}.

%In Figure~\ref{fig:arch_mask-head}, we present our multi-scale deformable mask head which is equally applicable for image segmentation and VIS.
In Figure~\ref{fig:arch_mask-head}, we present an overview of its architecture.
%
%In case of the latter, our encoder-decoder architecture outputs encoded backbone feature maps and $N = T * N_t$ object embeddings for clips with length $T$.
% Our encoder-decoder architecture outputs encoded backbone feature maps and $N = T * N_t$ object embeddings for clips with length $T$.
Our encoder-decoder architecture outputs encoded backbone feature maps and object embeddings per clip.
As shown in Figure~\ref{fig:method}, the mask head predicts binary instance masks by generating attention maps for each of the embeddings and applying a series of convolutions and upsampling operations.
Attention maps are generated by computing Multi-Head Attention (MHA), as in Equation~\ref{eq:mha}, between an embedding an the encoded feature maps of its corresponding frame in the clip.
At different scales of the upsampling process, corresponding backbone feature maps are added to improve segmentation performance.
The mask head predicts all instance masks in a clip at once and does not consider any explicit temporal information. 
In particular, each query only computes attention maps for its own frame as opposed to the deformable cross-attention computation in the decoder.

To take full advantage of the encoded multi-scale features, we propose the following design changes in order to improve training time and segmentation performance for VIS and image instance segmentation: 

\noindent \textbf{Training only positive matches.}
During the training of~\detr{}, the detection loss is computed via Hungarian matching between the ground truth and the object embeddings produced by the decoder.
We reduce the mask head training time by only computing masks and their corresponding losses embeddings which receive a positive matching with a ground truth object.
Since object queries are designed to exceed the number of objects per frame by a large margin, this results in a substantial reduction of training time.

\noindent \textbf{Encoded feature maps.}
The deformable attention allows us to encode all but the largest (H/4 x W/4) backbone feature maps.
While~\cite{DETR} only encodes the lowest (H/32 x W/32) feature map and hence must use raw backbone features in its mask head, we are able to add and upsample multiple Transformer-encoded feature maps as visualized in Figure~\ref{fig:arch_mask-head}.
%
% Following Deformable~\detr{}~\cite{deformable_detr}, our Transformer encoder-decoder only ignores the computational expensive largest backbone scale (H/4 x W/4). 
%
% Hence, we still add the original backbone feature map to the final upsampling step of our mask head.
%
The application of encoded feature maps allows for a direct connection between the mask head and Transformer encoder.

\noindent \textbf{Multi-scale attention maps.}
To obtain informative attention maps, the object embeddings must compute MHA with the feature maps in the preceding Transformer encoder-decoder.
Therefore, we are able to generate attention maps not only for a single but multiple scales and concatenate these at the corresponding stages of the upsampling process.
The original mask head only generates attention maps for the smallest scale (H/32 x W/32) and is not able to benefit from the additional connections to the object embeddings, \ie{}, the Transformer decoder, during the upsampling process.
It should be noted, that although our Transformer encoder-decoder computes deformable attention as in Equation~\ref{eq:tmsdmha}, the attention maps are generated via regular attention as in Equation~\ref{eq:mha}.

\noindent \textbf{MDC.}
Furthermore, we replace the convolutions in the mask head with Modulated Deformable Convolutions (MDC) ~\cite{zhu2018deformable}.
This not only boosts performance and convergence time but presents a more unified deformable approach. 

\noindent \textbf{End-to-end training.}
The additional connections of our mask head in to the preceding encoder-decoder blocks, see colored lines in Figure~\ref{fig:arch_mask-head}, result in a significant performance boost for an end-to-end training of the full model, \ie{}, including the backbone and encoder-decoder.
Without these connections, the authors of~\cite{DETR} did not observe any improvement for a similar end-to-end training.

% \tim{Find reason for why we do not use 3D convolutions like \vistr{}.}

% Note that we have not included the $3D$ convolutional mask head on top of it, as \vistr{} does.
% %
% We argue that, thanks to the temporal deformable attention, our queries share richer temporal features about the appearance of the object thus this module does not give us extra performance. 

\subsection{Multi-cue clip tracking}
For~\evis{} to run on sequences with arbitrary length, we apply a near-online clip tracking/stitching similar to~\cite{IFC,stem_seg,mask_prop}.
% , see Figure~\ref{fig:method}.
%
To this end, we sequentially match instance identities from consecutive and overlapping clips. 
%
% Given a trained~\evis{} model, the final runtime and performance can be modulated by adjusting the clip stride $S$, \ie, instance overlap, during inference.
%
% Up to date, \gui{drop up to date} the clip length that we have been using is $\tau=6$ frames, which is shorter than most of the videos in YouTube-VIS 2019. 
%
% Similar as other near-online video instance segmentation algorithms~\cite{IFC,stem_seg,mask_prop}, we match instances identities from two consecutive clips using predictions on the overlapping frames. 
%
% Let $\Upsilon$ and $\Upsilon_{\tau}$ be the set of track predictions of the already accumulated and next clip, respectively. 
The final set of tack predictions $\Upsilon$ is computed by matching the current $\Upsilon$ with the next $\Upsilon_k$ via the Hungarian algorithm~\cite{hungarian}.
%
% The matching of the current $\Upsilon$ with the next $\Upsilon_k$
% is performed via the Hungarian algorithm~\cite{hungarian}.
% based on mask Intersection over Union (IoU) costs.
%
% The cost matrix $C$ determines the optimal pair of indices $\hat{\sigma}$ that assign each of the instances from $\Upsilon$ with $\Upsilon_{\tau}$.
%
Instances in $\Upsilon_k$ without any match in $\Upsilon$ start a new instance identity.
%
% We follow~\cite{IFC} and improve track consistency by computing the volumetric soft IoU over consecutive sets of masks.
% $\text{SoftIoU}(m_i, m_j)$ between two masks $m_i, m_j$.
%
For tracking-by-detection/online methods it is common to perform the data association with multiple cues~\cite{Yang2019vis}.
We are the first to extend this idea to clip-level tracking and compute multiple cost terms for identities $i$ and $j$ in $\Upsilon$ and $\Upsilon_k$, respectively:

\begin{itemize}
    \item Mask cost via negative volumetric soft IoU~\cite{IFC} between consecutive overlapping sets of masks $m_{i,t}$ and $m_{j,t}$.
    \item Class cost $\mathcal{C}(c_i, c_j) = -1$ if $c_i = c_j$, and else 0, rewards consistent categories.
    % \item Class cost $\mathcal{C}(c_i, c_j) = 1$ if $c_i \neq c_j$ and otherwise 0 to reward consistent categories.
    \item Score cost $\mathcal{S}(s_i, s_j)=|s_i - s_j|$ matches clips with similar confidence.
    % \item Class score cost $\mathcal{S}(s_i, s_j)=(s_i + s_j) / 2$ encourages matches of clips with high confidence.
    % \item Bounding box center cost $\mathcal{B}(b_i,b_j)=||b_{i\{c_x,c_y\}}-b_{j\{c_x,c_y\}}||$ potentially allows matching for occluded frames in which no mask is available.
\end{itemize}

The contribution of each cue is controlled by their corresponding weights $\sigma_{mask}$, $\sigma_{class}$ and $\sigma_{score}$. % and $\sigma_{bbox}$.
In cases of strong occlusion and imperfect mask predictions, the identity matching can additionally rely on consistent class and score information which further improves the identity preservation across a sequence.

\section{Experiments}
This section provides the most relevant details of our implementation, and the experimental setup for the following ablation studies and benchmark evaluations.
If not otherwise specified all results in the paper are obtained with a ResNet-50~\cite{resnet} backbone and follow the hyperparameters of~\cite{deformable_detr}.
For additional implementation and training details we refer to the appendix.

% \subsection{Implementation details}

\noindent \textbf{Multi-scale deformable mask head.}
We train our mask head jointly with a pre-trained Deformable~\detr{}~\cite{deformable_detr} on COCO~\cite{COCO} for 24 epochs, decaying the learning by $0.1$ after epoch 15.
For batch size 2, we use 8 GPUs with 32GB memory for 2 days (345 GPU hours).
The initial learning rates of the backbone, encoder-decoder, and mask head are $1e^{-5}$, $2e^{-5}$, and $2e^{-4}$, respectively. 
%
% The remaining hyperparameters follow the default configuration from~\cite{deformable_detr}.

\noindent\textbf{\evis{}.}
We initialize our~\evis{} model from the preceding end-to-end instance mask head training and then fine-tune for additional 10 epochs on the respective VIS dataset.
With one clip per GPU, we use 4 GPUs for 1.5 days (120 GPU hours) with 18GB memory for YouTube-VIS 2019~\cite{Yang2019vis} dataset.
The increased number of objects per frame in both YouTube-VIS 2021~\cite{Yang2019vis} and OVIS~\cite{ovis} require 24GB of memory.
%
% We obtained a substantial performance increase from applying the same incremental weighting of the 6 Transformer decoder auxiliary loss terms as in~\cite{AuxLoss}.
%
In contrast to~\cite{vistr,IFC}, we are able to train with data augmentations on different input scales and apply a learned additive temporal encoding.
If not otherwise specified, all models use clip size $\tau=6$, stride $S=4$ and reference point keys $K_{curr}=K_{temp}=4$.
We run multi-cue clip tracking with $\sigma_{mask} = \sigma_{class} = \sigma_{score} = 1$. 
For the ablations, we report best validation scores after training 10 epochs.
%
% \todo{Specify topk=10 for ablations, and then for experiments/final results topk=20 or topk=30}

% We set $M=8$, $L=4$, $K_{curr}=4$ and $K_{temp}=4$ for the temporal deformable attention module. 
%
% We use AdamW \cite{AdamW} optimizer with initial learning rate of $10^-4$, $\beta_1 = 0.9$, $\beta_1 = 0.999$ and weight decay $10^-4$. 
%

\noindent \textbf{Datasets and metrics.}
We evaluate results on the \textit{YouTube-VIS 2019/2021}~\cite{Yang2019vis} datasets which contain 2883 and 3859 high quality videos with 40 unique object categories, respectively.
The latter provides improved ground truth annotations.
The OVIS~\cite{ovis} dataset includes severe occlusion scenarios on 901 sequences.
We measure the Average Precision (AP) as well as Average Recall (AR).
For instance mask prediction on images, we report the AP based on mask IoU on the common COCO~\cite{COCO} dataset.

\begin{table*}[t]

\caption{
Ablation of our main~\textbf{\evis{} contributions} and the incremental built from a naive Deformable~\vistr{} to our final model.
\textit{Increase spatial inputs} denotes multi-scale training on higher input resolutions.
% All experiments use encoder $K_2=2$ and decoder $K_2=1$.
%
% We use the exact same data transformation as VisTr.
%
}

\label{tab:ablation_main}

\centering

\resizebox{\columnwidth}{!}{%

\begin{tabular}{lrccc | cccHc}

\toprule

%&\thead{Loss \\ weighting} &\thead{New matcher \\ loss}
% 
%&\thead{Multi-scale \\ mask head}
%&\thead{Spatial \\ adjustments}

Method  &  \thead{Clip \\ size $\tau$}   & $K_{curr}$ & $K_{temp}$  &\thead{Feature \\ scales} & AP & $\Delta$ AP &\thead{Training \\ GPU hours} & FPS &\thead{\#params}\\

\toprule
\multirow{2}*{\shortstack[l]{\vistr{}}}

& 36  & All  & All   & 1 & 36.1 & -- & 350 & 69.9 &57M\\
%& 36  & All  & All   & 4 & -- & -- & OOM? &-- & --\\
& 6  & All  & All   & 4 & -- & -- &OOM & X &64M\\
%\vistr  & 36  & All  & All   & 1 & 36.1 & -- & 350 & 69.9 &57M\\

\midrule
\multirow{4}*{\shortstack[l]{Deformable \\ \vistr{}}}       
                    
                    &36  &4  &4    &1   &34.2  & -- &260 &64.3 &47M\\
                    
                    % &36  &4  &4    &4  & --     & --   & -- & -- &68M\\
                    
                    &36  & 4  & 0  &1   &35.3   & -- &155 &83.2 &36M\\
                    
%                    &36  & 4  & 0  &4   &--   & -- &XX &XX &XX\\
                    
%                     &36  & 4  & 0  &4   &32.2   & -- &167 &XX &41.9M\\

                    &6  &4  &4   & 1 &   34.0   & -- &48  &31.8  &41M\\
                    
                    &6  &4  &0   & 1 &  32.4    & -- &30  &XX  &36M\\

\midrule
\evis{}                             &  &  &     & &   &  && &\\
% \evis{}                             &6  &6  &6     &60 Object &33.9   &63  &13.3 &X\\
+ Increase spatial inputs          &6  &4  &4      & 4 & 35.9  & +1.9  & 93    & 25.5  & 48M\\
+ Instance-aware object queries     &6  &4  &4      & 4 & 37.0  & +1.1  & 112    & 22.3  & 48M\\
+ Multi-scale mask head             &6  &4  &4      & 4 & 40.2  & +3.2  & 120      & 23.4  & 48M\\
+ Multi-cue clip tracking           &6  &4  &4      & 4 & 41.9  & +1.7  & 120      & 18.4  & 48M\\
% + Auxiliary loss weighting, loss all frames  &6  &4  &4  & 4 &44.2  & +2.9  & 3:15hx4   & X  & 48M\\
+ Auxiliary loss weighting  &6  &4  &4  & 4 &44.0  & +2.1  & 120   & 18.4  & 48M\\

\bottomrule

\end{tabular}
}

\vspace{-0.5cm}

\end{table*}

\subsection{Ablation studies}

We present ablations demonstrating the effectiveness of our contributions for~\evis{} and provide detailed insights on the new mask head and clip tracking.

\noindent \textbf{Making~\evis{} work for VIS.}
%
% \todo{1) New ablation 6-4-0-1 reasoning: different from full clip size, with clip size 6 temporal connections bring performance. 2) VisTR 6-All-All-4: Impossible to train VisTR with 4 Feature scales, motivates use of deformable attention in order to go multi-scale and benefit from improved mask head.}
In Table~\ref{tab:ablation_main}, we demonstrate the shortcomings of a naive deformabilization, \ie, replacement of~\vistr{}'s~\cite{vistr} full attention with deformable attention.
We indicate full attention over all pixels wit $K_{temp}=\text{All}$.
Running temporal deformable attention offline, \ie, with clip size $\tau=36$, reduces training time compared to~\vistr{}, but converges with an unsatisfactory performance of 34.2.
This is due to the incapability of spatially restricted reference points to compute meaningful temporal attention connections over large clip sizes.
%
% The learnable reference points and attention weights furthermore result in an increase of training parameters to 68M.
%
This assumption is supported by the 1.1 improvement of an offline Deformable~\vistr{} without any temporal connections, \ie, $K_{temp}=0$.
A reduction of the clip size to $\tau=6$ in a near-online fashion mitigates the reference point issues while also requiring only a fraction of the original training time (155 vs. 48 GPU hours).
To further support our hypothesis, we demonstrate how removing temporal connections $K_{temp}=0$ for smaller clip sizes $\tau=6$ does indeed deteriorate the performance.
However, without fully capitalizing on the efficiency of deformable attention its best version with 34.0 is still inferior to~\vistr{}.

The first~\evis{} row demonstrates the potential gains (1.9) from increasing the number of feature scales to $L=4$ and training on higher input resolutions.
% with multi-scale data augmentations
%
The transition to multiple scales is a prerequisite for the application of our new mask head and only feasible to train on smaller clip sizes.
The same model with full attention (second row) results in out-of-memory (OOM) errors.
Naturally, this increases the total number of parameters and training time but both remain far below~\vistr{}.
Instance-aware object queries come with a neglectable increase in training time but provide a 1.1 AP boost.
The individual contributions of the mask head and clip tracking additions are ablated in Table~\ref{tab:ablation_mask_head} and~\ref{tab:ablation_multi_cues}, respectively.
Both result in additional performance boosts without substantially increasing the computational costs.
A cascaded weighting of the decoder auxiliary loss terms as applied in~\cite{AuxLoss} increases results further.
Our final model benefits from deformable attention with low training times and parameter counts surpassing a naive Deformable~\vistr{} approach by 11.1 points.

In Table~\ref{tab:ablation_temporal}, we ablate different clip sizes $\tau$ and number of temporal sampling keys $K_{temp}$ for our final configuration.
We used the optimal clip size $\tau=6$ for our~\evis{} ablations in Table~\ref{tab:ablation_main}.
Both smaller and larger clip sizes resulted in worse performance either due to the lack of temporal connections or the aforementioned problem of spatially local reference points in the encoder, respectively.
Running all $L=4$ feature scales was only possible for a clip length of up to $\tau=12$.
Furthermore, we ablate the removal of temporal connections which resulted in a large relative drop for the challenging OVIS~\cite{ovis} dataset.

\begin{table*}[t]
\parbox[t][][t]{.48\linewidth}{

\centering

\caption{%
    Ablation for the \textbf{multi-scale mask head} on COCO~\cite{COCO}.
    The baseline applies the original mask head as in~\detr{} with Deformable~\detr{}~\cite{deformable_detr}.
    %
    % We then show how each newly added component helps improving the performance.
}
\label{tab:ablation_mask_head}

\resizebox{0.48\columnwidth}{!}{%
\begin{tabular}{l | cHHcH}

\toprule
Mask head & \thead{Mask \\ mAP} &AP$_{50}$ &AP$_{75}$ & \thead{Training \\ GPU hours} & FPS\\

\midrule

% ~\detr{}~\cite{DETR}         &$33.3$ & & & -- & --\\

% \midrule

Baseline as in~\cite{DETR} with~\cite{deformable_detr}          &$23.7$ & & &242 & X\\
+ Train only positive matches       & $24.5$ &$46.8$ &$23.1$ &58 & X \\
+ Encoded feature maps              & $25.0$ &$47.3$ &$23.9$ &56  & X\\
+ Multi-scale attention maps        & $29.2$ &$53.8$ &$28.3$ &59  & X\\
+ MDC                               & $31.1$ &$54.9$ &$31.1$ &78  & X\\
+ End-to-end training                  & $38.0$ &$61.4$ &$40.1$ &345  & X\\
% + Finetuning without aux loss           & $36.2$ & & & \\
% + Auxilliary loss       & $37.5$ & & & \\

\bottomrule

\end{tabular}
}
}\hfill
\parbox[t][][t]{.48\linewidth}{
\centering

\caption{
Removing \textbf{temporal connections} with $K_{temp}=0$ results in performance drops across all datasets.
We observe an optimal clip size of $\tau=6$.
%
%  We evaluate the impact of the temporal connections on the multi-scale deformable attention on each dataset.
%
%  The benefit obtained increases as the dataset difficulty also increases
%
%  \todo{rerun clip size 3 9 12 in final config}.
}
 
 \label{tab:ablation_temporal}
 
% \vspace{0.17cm}

\resizebox{0.48\columnwidth}{!}{%

\begin{tabular}{cc|cc|cc|cc}
  \toprule
    \multirow{2}*{\thead{Clip \\ size $\tau$}}& \multirow{2}*{$K_{temp}$} & \multicolumn{2}{c|}{YT-VIS 19~\cite{Yang2019vis} } & \multicolumn{2}{c|}{YT-VIS 21~\cite{Yang2019vis}} & \multicolumn{2}{c}{OVIS~\cite{ovis}} \\
    
    \cmidrule{3-8}
    
    %  & AP & AP$_{75}$ & AP$_{50}$ & AP & AP$_{75}$ & AP$_{50}$ & AP & AP$_{75}$ & AP$_{50}$ \\
%   \midrule
%  0  &$41.2$ &$45.4$  &$63.8$ &$39.5$ &$42.7$  &$61.7$ &$19.7$ &$18.8$  &$39.3$ \\
%   4  &$44.1$ &$48.1$  &$65.1$ &$41.9$ &$46.0$  &$64.8$ &$23.2$ &$21.7$ &$44.4$\\
  
  & & AP & $\Delta$ AP & AP & $\Delta$ AP & AP & $\Delta$ AP \\
  \midrule
  6 & 4 & $44.4$ & -- &$43.1$ & -- &$23.8$ & --\\
  \midrule
  6 & 0 & $41.2$ & -3.2 &$39.5$ & -3.6 &$19.7$ & -4.1 \\
  3 & 4 & 41.0 & -3.4 &-- & -- &-- & --\\
  9  & 4 & 42.4 & -2.0 &-- & -- &-- & --\\
  12  & 4 & 41.6 & -2.8 &-- & -- &-- & --\\
  \bottomrule
 \end{tabular}
}
}

\vspace{-0.3cm}

\end{table*}

\noindent \textbf{Multi-scale mask head.}
We evaluate our contributions on the mask head in Table~\ref{tab:ablation_mask_head} on COCO~\cite{COCO} instance segmentation.
The baseline represents a straightforward application of the original~\detr{}~\cite{DETR} mask head with the Deformable~\detr{}~\cite{deformable_detr} detector, which is not only 9.6 points worse than~\cite{DETR} (see Table~\ref{tab:eval_COCO}), but suffers from an unfeasible long training time largely due to the increased number of object queries in~\cite{deformable_detr}.
By computing instance masks only for queries positively matched with a ground truth object, we are able to reduce the training time 4-fold.
The following two additions take full advantage of the encoded multi-scale features of~\cite{deformable_detr} and result in a mask AP of 29.2.
For top performance, we further add MDC~\cite{zhu2018deformable} and train the entire model end-to-end.
Our mask head without end-to-end training is still inferior to~\detr{}.
This can be attributed to the sparse computation of deformable attention which makes the generated attention maps less suitable for pixel-level segmentation.
However, the end-to-end training fully realizes the potential of the additional connections between our mask head and the encoder-decoder.
% thereby mitigating these aforementioned issues.
%
The increased training time is justified by the overall 4.7 point improvement over~\detr{}.

% We show how we have built our mask head architecture on Table ~\ref{tab:ablation_mask_head}. 
% %
% We demonstrate that the mask head designed for \detr{} suffers when plugged in on Deformable \detr{}. 
% %
% Our solution effectively leverages multi-scale features added with the later to boost the performance.  

% \input{tables/ablation_multi_cues}

\begin{table*}[t]
\parbox[t][][t]{.48\linewidth}{

\centering

\caption{
Comparison of instance segmentation results on~\textbf{COCO}~\cite{COCO}.
\mbox{Mask R-CNN} is from detectron2~\cite{wu2019detectron2}.
}
\label{tab:eval_COCO}

\resizebox{0.48\columnwidth}{!}{%
\begin{tabular}{l | cccccc|c}
\toprule
Methods &AP &AP$_{50}$ &AP$_{75}$  &AP$_l$  &AP$_m$  &AP$_s$  &FPS\\
\midrule

\detr{}~\cite{DETR}                &$33.3$ &$56.5$ &$33.9$ &$53.1$ &$36.8$ &$13.5$ &--\\
IFC~\cite{IFC}                 &$35.1$ & --      & -- &-- &-- &-- &--\\
Mask R-CNN~\cite{he2017mask}      &$37.2$ &$58.5$ &$39.8$ &$53.3$ &$39.4$ &$18.6$ &21.4\\
Mask2Former~\cite{Mask2Former}  &$43.7$ &-- &-- &$64.8$ &$47.2$ &$23.4$ &13.5\\
\midrule
\textbf{Ours}                &$38.0$ &$61.4$ &$40.1$ &$59.8$ &$41.4$
&$17.9$ &12.1\\

\bottomrule
\end{tabular}

}
}\hfill
\parbox[t][][t]{.48\linewidth}{
\centering

\caption{
%
% Impact of each component from our multi-cue clip tracking/stitching strategy.
%
% YouTube-VIS 2019~\cite{Yang2019vis}
Contribution of the additional class and score cost terms in our \mbox{\textbf{multi-cue}} \textbf{clip tracking}.
}
 \label{tab:ablation_multi_cues}

\resizebox{0.48\columnwidth}{!}{%

\begin{tabular}{l|ccc}
  \toprule
    Clip tracking cues  & AP &AP$_{50}$ &AP$_{75}$\\
  \midrule
  %\multirow{2}*{Single-cue}
   %&Mask IoU &$39.9$ &$61.3$ &$43.8$  \\
  Vol. soft mask IoU &$42.1$ &$63.2$ &$46.7$  \\
  %\midrule
  %\multirow{2}*{Multi-cue}
  Vol. soft mask IoU + Score &$42.8$ &$65.1$ &$46.9$   \\
  Vol. soft mask IoU + Class &$43.1$ &$66.0$  &$47.2$   \\
  Vol. soft mask IoU + Score + Class &$44.4$ &$66.8$  &$48.5$  \\

  \toprule
 \end{tabular}
}
}

\vspace{-0.2cm}

\end{table*}

\noindent \textbf{Multi-cue clip tracking.}
To improve the track consistency between clips, we introduce additional cues to the common mask-based clip tracking.
The clip tracking row in Table~\ref{tab:ablation_main} replaces the original~\vistr{} mask IoU cost term with a combination of volumetric soft mask IoU, class, and score costs.
After tuning the cost weighting parameters, we ablate their individual contributions in Table~\ref{tab:ablation_multi_cues}.
The new class and score terms provide an overall boost of 2.3 AP points.

\begin{table*}[t]

\caption{
Comparison of VIS methods on the~\textbf{YouTube-VIS 2019/2021}~\cite{Yang2019vis} validation sets.
%
% $^{\dagger}$ indicates ResNet-50-DCN .
%
FPS measurements denoted with $^{\ast}$ are extracted from~\cite{VISOLO}.
With $^{\ast\ast}$ and $^{\dagger}$the  we denote a joint training with COCO~\cite{COCO} and unpublished methods, respectively.
}

\vspace{-0.5cm}

\label{tab:eval_vis_all}
\begin{center}
\resizebox{\columnwidth}{!}{%
\begin{tabular}{l l|c|r|ccccc|ccccc|ccc}
\toprule
\multicolumn{2}{c|}{\multirow{2}*{Method}} & \multicolumn{1}{c|}{\multirow{2}*{Backbone}} & \multicolumn{6}{c}{YT-VIS 19~\cite{Yang2019vis}} & \multicolumn{5}{|c}{YT-VIS 21~\cite{Yang2019vis}} & \multicolumn{3}{|c}{OVIS~\cite{ovis}}\\
\cmidrule{4-17}
& &  & \multicolumn{1}{c|}{FPS} & AP & AP$_{50}$ & AP$_{75}$ & AR$_1$ & AR$_{10}$ & AP & AP$_{50}$ & AP$_{75}$ & AR$_1$ & AR$_{10}$ & AP & AP$_{50}$ & AP$_{75}$\\

\midrule
\multirow{7}*{\rotatebox[origin=c]{90}{Online}} & MaskTrack-RCNN~\cite{Yang2019vis} & R50 & $^{\ast}$26.1 & 30.3 & 51.1 & 32.6 & 31.0 & 35.5 &  28.6 & 48.9 & 29.6 & 26.5 & 33.8 & $15.4$ & $33.9$ & $13.1$ \\
& SipMask~\cite{sip_mask} & R50 & $^{\ast}$35.5 & 33.7 & 54.1 & 35.8 & 35.4 & 40.1  & 31.7 & 52.5 & 34.0 & 30.8 & 37.8 & $14.3$ & $29.9$ & $12.5$ \\
& SG-Net~\cite{sg_net} & R50 & $^{\ast}$23.0 & 34.8 & 56.1 & 36.8 & 35.8 & 40.8 & -- & -- & -- & -- & -- & -- & -- & --\\
& CompFeat~\cite{CompFeat} & R50 & $-$ & 35.3 & 56.0 & 38.6 & 33.1 & 40.3 & -- & -- & -- & -- & -- & -- & -- & --\\
& CrossVIS~\cite{cross_vis} & R50 & 39.8 & 36.3 & 56.8 & 38.9 & 35.6 & 40.7 & 34.2 & 54.4 & 37.9 & 30.4 & 38.2 & $18.1$ & $35.5$ & $16.9$ \\
& STMask~\cite{stmask} & R50-DCN & 28.6 & 33.5 & 52.1 & 36.9 & 31.1 & 39.2 & 30.6 & 49.4 & 32.0 & 26.4 & 36.0 & -- & -- & --\\
& VISOLO\cite{VISOLO} &50 & 40.0 & 38.6 & 56.3 & 43.7 & 35.7 & 42.5 & 36.9 & 54.7 & 40.2 & 30.6 & 40.9 & -- & -- & --\\

\midrule
\multirow{8}*{\rotatebox[origin=c]{90}{Offline}}
& VisTR~\cite{vistr} & R50 & 69.9  & 36.2 & 59.8 & 36.9 & 37.2 & 42.4 & -- & -- & -- & -- & -- & -- & -- & --\\
& SeqMask-RCNN~\cite{prop_reduce} & R50 & $^{\ast}$3.8 & 40.4 & 63.0 & 43.8 & 41.1 & 49.7 & -- & -- & -- & -- & -- & -- & -- & --\\
& IFC~\cite{IFC} & R50 & \textbf{107.1} & 41.2 & 65.1 & 44.6 & 42.3 & 49.6 & 35.2 & 57.2 & 37.5 &-- &-- & -- & -- & --\\

& IFC~\cite{IFC} & R101 & 89.4 & 42.6 & 66.6 & 46.3 & 43.5 & 51.4 & -- & -- & -- &-- &-- & -- & -- & --\\
& SeqFormer$^{\dagger}$~\cite{SeqFormer} & R50 & 12 & 45.1 & 66.9 & 50.5 & 45.6 & 54.6 & 40.5 & 62.4 &43.7 &36.1 & 48.1 & -- & --\\
& SeqFormer$^{\dagger\ast\ast}$~\cite{SeqFormer} & Swin-L & -- & 59.3 & 82.1 & 66.4 & 51.7 & 64.4 & 51.8 & 74.6 &58.2 &42.8 & 58.1 & -- & --\\
% BackB FPS  AP & AP$_{50}$ & AP$_{75}$ & AR$_1$ & AR$_{10}$ & AP & AP$_{50}$ & AP$_{75}$ & AR$_1$ & AR$_{10}$ & AP & AP$_{50}$ & AP$_{75}$

& Mask2Former$^{\dagger}$~\cite{Mask2Former} &R50 &-- & \textbf{46.4} & \textbf{68.0} & \textbf{50.5} &-- & -- & 40.6  & 60.9 &41.8 &-- & -- &-- &-- & --\\
%& Mask2Former &R101 &-- & \textbf{49.4} & \textbf{72.8} & \textbf{54.2} &-- & -- & 42.4  & 65.9 &45.8 &-- & -- &-- &-- & --\\
& Mask2Former$^{\dagger}$~\cite{Mask2Former} &Swin-L &-- & \textbf{60.4} & \textbf{84.4} & \textbf{67.0} &-- & -- & 52.6  & 76.4 & 57.2 &-- & -- &-- &-- & --\\

\midrule
\multirow{5}*{\rotatebox[origin=c]{90}{Near-online}}
& MaskProp~\cite{mask_prop} (T=12) & R50 & -- & 40.0 & -- & 42.9 & -- & -- & -- & -- & -- & -- & -- & -- & -- & --\\
& STEm-Seg~\cite{stem_seg} & R50 & $^{\ast}$3.0 & 34.6 & 55.8 & 37.9 & 34.4 & 41.6 & -- & -- & -- & -- & -- & $13.8$ & $32.1$ & $11.9$ \\
& IFC~\cite{IFC} (T=5, S=1) & R50 & 46.5 & 39.0 &60.4 &42.7 & 41.7 & 51.6 & -- & -- & -- & -- & -- & -- & -- & --\\
& \textbf{\evis{}} (T=6, S=4) & R50 & 18.4  &44.4 & 66.7 &48.6 &42.4 &51.6 &\textbf{43.1} & \textbf{66.8} & \textbf{46.6} & \textbf{38.0} & \textbf{50.1} &\textbf{23.8} &\textbf{48.0} &\textbf{20.8}\\
%& \textbf{\evis{}} (T=6, S=4) &R101 & 18.4  &-- &-- &-- &-- &-- &\textbf{44.4} & \textbf{68.8} & \textbf{48.3} & \textbf{38.8} & \textbf{51.2} &\textbf{X} &\textbf{X} &\textbf{X}\\
& \textbf{\evis{}} (T=6, S=4) &SwinL & 18.4  &57.1 & 80.8 & 66.3 &50.8 &61.0 & \textbf{54.4} & \textbf{77.7} & \textbf{59.8} & \textbf{43.8} & \textbf{57.8} &\textbf{34.6} &\textbf{58.7} &\textbf{36.8}\\

% & \textbf{\evis{}} (T=6, S=4) & 101 & XX &44.7 &67.9 &49.5 &43.4 &51.8  & -- & -- & -- & -- & -- & -- & -- & --\\%

%
% to make "Near-online" fit in the cell
%
\noalign{\vskip 2mm}    

\bottomrule
\end{tabular}}
\end{center}

\vspace{-0.8cm}

\end{table*}

\subsection{Benchmark evaluation}
\vspace{-0.2cm}
\noindent \textbf{Video instance segmentation.}
To demonstrate the effectiveness of~\evis{} in comparison with other VIS methods, we report results on the YouTube-VIS 2019 and 2021~\cite{Yang2019vis} dataset in Table~\ref{tab:eval_vis_all}.
Our method achieves state-of-the-art performance with respect to all previous methods on the YouTube-VIS 2021 dataset by significant margins of 2.5 and 1.8 for ResNet-50~\cite{resnet} and Swin-L~\cite{SwinTransformer} backbones, respectively.
The benefits of our temporal connections and multi-scale feature encoding in conjunction with the mask head are most apparent by our 7.9 improvement over IFC~\cite{IFC}.
Both ~\vistr{} and IFC achieve lower runtimes mirroring the relation between Deformable~\detr{}~\cite{deformable_detr} and~\detr{}~\cite{DETR}.
Furthermore, both methods rely on an expensively pretrained~\detr{} limiting their adaptability dramatically.
In addition to YouTube-VIS, we are the first Transformer-based method to present results on the OVIS~\cite{ovis} dataset.
We surpass the previous best method~\cite{cross_vis} by 5.7 points for ResNet-50.
%
%Note, this methods obtains a comparable performance to~\vistr{} on YouTube-VIS.
%
Due to its pixel-level encoding of image features,~\evis{} excels on OVIS' challenging occlusions scenarios.

\noindent \textbf{Image instance segmentation.}
The new multi-scale mask head does not only boost VIS performance but also excels on image segmentation.
In Table~\ref{tab:eval_COCO}, we evaluate performance on COCO~\cite{COCO} even surpassing Mask R-CNN~\cite{he2017mask}.
We present this as contribution beyond the VIS community and regard our mask head as a valuable completion of Deformable~\detr{}. 
Interestingly,~\evis{} is superior to Mask2Former despite the inferiority of our mask head for image segmentation.
We regard this as a strong sentiment for our~\evis{} video approach.

\vspace{-0.2cm}
\vspace{-0.2cm}

\section{Conclusion}
\vspace{-0.2cm}
We have proposed a novel VIS method which applies a Transformer encoder-decoder architecture to clips of frames in a near-online fashion.
To mitigate the efficiency issues of previous Transformer-based methods suffering from quadratic input complexity, we propose temporal deformable attention with instance-aware object queries.
Deformable attention allows us to benefit from multi-scale feature maps and led to the introduction of a new powerful instance mask head.
% for object query embeddings.
%
Furthermore, we present multi-cue tracking with class and score terms.
Our~\evis{} method achieves state-of-the-art results on two VIS datasets and hopefully paves the way for future applications of deformable attention for VIS.

\clearpage

\ifarxiv
    \appendix

    \def\suppabstract{This section provides additional material for the main paper:
    \S\ref{sec:imp_details_appendix} contains further implementation details for our \evis{} method and its training. 
    In \S\ref{sec:ablation_appendix}, we discuss further ablations on the effect of the clip size and temporal sampling points.
    Furthermore, we complement the qualitative results of the main paper with selected illustrations in (\S\ref{sec:qual_results_appendix}).
    These include qualitative results, more detailed attention maps and failure cases.
    }

    \section*{Appendix}
    \suppabstract
    
    \newcommand{\sref}[1]{Sec.~\ref{#1}}
    \setcounter{table}{0}
    \renewcommand{\thetable}{A.\arabic{table}}
    
    \setcounter{figure}{0}
    \renewcommand{\thefigure}{A.\arabic{figure}}
    
    % \begin{abstract}

% small abtract

% \end{abstract}

\section{Implementation details}

\label{sec:imp_details_appendix}

\noindent \textbf{Multi-scale mask head}
To improve convergence of the end-to-end full model training, we increase the dice and mask loss weights to $\lambda_{DICE}=\lambda_{MASK}=8$.
Furthermore, we add both loss terms to the auxiliary losses of the \nth{3} decoder layer.
For trainings of the new mask head with~\evis{}, we keep the original mask loss weights but also add the corresponding terms to the \nth{3} auxiliary loss.
To further speed up the inference, we reduce the \textit{top k} from 100 as in ~\cite{deformable_detr} to 50.

\noindent \textbf{\evis{}}
We train our model with a total of 60 object queries for YouTube-VIS 2019 and 180 object queries for YouTube-VIS 2021 and OVIS.
With a clip size of $\tau=6$ this assigns 10 or 30 queries to each frame.
OVIS includes sequences with up to 44 unique instances and hence requires an increased number of object queries.
The class head is solely responsible for categorization of each object query.
% the predicts class scores for each output query.
%
This means, the mask head predicts a single mask for each object query and does not produce per-category outputs as Mask R-CNN~\cite{he2017mask}.
In order to associate a single class with a trajectory of queries, we compute the mean score over each class.
This results in a total of number of classes times number of object queries per frame trajectories.
The final output is selected in a top-k manner from the total set of trajectories.
It should be noted, if k is larger than the number of queries per frame, a single trajectory is associated with multiple labels.
To work as similar as possible to~\vistr{}~\cite{vistr}, our ablation experiments apply $k=10$.
For our benchmark experiments, we increase the value to $k=20$ and $k=30$ for YouTube-VIS 2019 and YouTube-VIS 2019/OVIS, respectively.

%}
%
For the auxiliary loss weighting~\cite{AuxLoss}, we use the following incremental weights from the first to the final layer: $1/2$, $5/30$, $4/30$, $3/30$, $2/30$ and $1/30$.
%The incremental auxiliary loss term weighting applies the same weights to each decoder layer as in ~\cite{AuxLoss}: $1/2$, $5/30$, $4/30$, $3/30$, $2/30$ and $1/30$.
%
Such a weighting decreases the influence of early decoder layers with an emphasized focus on optimizing the outputs of the final layer.
In contrast to~\cite{AuxLoss}, we keep the same weighting during the entire training.
However, the weighting is not applied to the mask auxiliary loss on the third layer.
As the total contribution from non-mask losses is heavily reduced, we use $\lambda_{DICE}=\lambda_{MASK}=1$, since we did not observe any benefit in this case from increasing its contribution.  
Furthermore, we use $\lambda_{CLASS} = 1$ weight for the class cost and matching, following~\cite{vistr}. 
In comparison to~\cite{deformable_detr}, our encoder-decoder model introduces only changes to the dimensions of a few parameters, namely, object queries and linear projections for sample offset and attention weight.
We train the object queries, the introduced temporal learned embedding and the classification head from scratch. 
For the linear projections, we duplicate the existing pre-trained weights for each new temporal position, adapted to the number of points $K_{temp}$. 
We apply initial learning rates of $1e^{-5}$ and $1e^{-4}$ for the backbone, and rest of the model including the mask head, respectively. 
We drop these by $0.1$ at epoch $3$ and $7$ for YouTube-VIS 2019, $4$ and $8$ for YouTube-VIS 2021 and $6$ and $10$ for OVIS.

\noindent \textbf{Learnable sample offset and attention weight parameters}
To circumvent the quadratic input complexity of regular attention, deformable attention~\cite{deformable_detr} learns linear projections which infer the sample offsets $\Delta\vp_{mqk}$ and attention weights $\emA_{mqk}$ for a query $q$, sampling point $k$ and attention head $m$.
In our temporal deformable attention formulation, we separately learn linear projections for the current and temporal frames, see Table~\ref{tab:learn_offset_and_weight_params_dims} for their dimensions.
This allows to individually set $K_{curr}$ and $K_{temp}$ and ablate configurations without temporal connections.
Given a clip with $\tau=6$, the queries assigned to a frame $t$ apply their $\tau - 1$ learned temporal projections to the remaining frames in the following frame order:

We follow VIS convention to run inference on reduced input resolutions with 360 pixels but upsample to the required benchmark resolution before the clip tracking.
The runtime frames per second (FPS) measurement do not include this upsampling.

\begin{table}
\centering
\begin{tabular}{c | c}
    Query frame $t$ & Temporal frame indices \\
    \midrule
    0 & [1, 2, 3, 4, 5] \\
    1 & [0, 2, 3, 4, 5] \\
    2 & [0, 1, 3, 4, 5] \\
    3 & [0, 1, 2, 4, 5] \\
    4 & [0, 1, 2, 3, 5] \\
    5 & [0, 1, 2, 3, 4]
\end{tabular}

\end{table}
% \begin{itemize}[leftmargin=2.0in]
%     \item [Frame $t=0$] [1,2,3,4,5]
%     \item [Frame $t=1$] [0,2,3,4,5]
%     \item [Frame $t=2$] [0,1,3,4,5]
%     \item [Frame $t=3$] [0,1,2,4,5]
%     \item [Frame $t=4$] [0,1,2,3,5]
%     \item [Frame $t=5$] [0,1,2,3,4]
% \end{itemize}

This requires the same projection parameters to predict offsets/weights with different temporal distances to the current frame which is possible due to the learned temporal encoding added to each query.
Furthermore, the explicit discrimination between the current and temporal frames with respect to a query allows decoder object queries to focus on predictions for their frame while taking additional temporal information under consideration.
It should also be noted that all the terms in Table~\ref{tab:learn_offset_and_weight_params_dims} scale linearly with the number of frames in a clip and and do not depend on the input resolution.

\begin{table*}
\centering
\caption{
    Dimensions of learnable sample offset and attention weight parameters with object queries $N$, attention heads $m$, feature scales $L$ and clip size $\tau$.
    }

\label{tab:learn_offset_and_weight_params_dims}

\begin{tabular}{r | c | c}
\toprule
                    & Current frame & Temporal frames \\
\midrule

Attention weights   & $N \times M \times L \times K_{curr} \times 1 \times 1$ & $N \times m \times L \times K_{temp} \times \tau -1 \times 1$  \\
Sampling offsets    & $N \times M \times L \times K_{curr} \times 1 \times 2$ & $N \times m \times L \times K_{temp} \times \tau -1 \times 2$ \\

\bottomrule
\end{tabular}
\end{table*}

\section{Ablation studies}

\label{sec:ablation_appendix}

% In Table~\ref{tab:ablation_temporal_extended}, we extend the analysis of the main paper on the effect of adding more or less temporal information by altering the clip stride $S$ and number of $K_{temp}$ for the encoder and decoder separately.
In Table~\ref{tab:ablation_temporal_extended}, we extend the analysis of the main paper on the effect of adding more or less temporal information by altering the number of $K_{temp}$ for the encoder and decoder separately.
%
% In order to run different strides $S$, we apply the same model with a different clip tracking.
%
% In contrast, t
The ablation of different $K_{temp}$ require individually trained models with differing number of parameters.
Given a trained~\evis{} model, the final runtime and performance can be modulated by adjusting the clip stride $S$, \ie, instance overlap, during inference.
For experiments with differing clip size $\tau$, we adjust the stride to keep the number of overlapping frames between constant.
We set $K_{curr}=4$ as in~\cite{deformable_detr} for all experiments and obtain the same optimal value for $K_{temp}$.

% $K_{temp}=4$ is the default from~\cite{deformable_detr}.

% \todo{maybe: discriminate $K_{temp}$ for encoder and decoder. but for $K_{temp}=4$}

\begin{table*}[t]

 \caption{
Removing \textbf{temporal connections} with $K_{temp}=0$ results in performance drops across all datasets.
We observe an optimal clip size of $\tau=6$.
%
%  We evaluate the impact of the temporal connections on the multi-scale deformable attention on each dataset.
%
%  The benefit obtained increases as the dataset difficulty also increases
%
%  \todo{rerun clip size 3 9 12 in final config}.
}
 
 \label{tab:ablation_temporal_extended}
 
  \center
  \begin{tabular}{cHc|Hcc|cc|cc}
  \toprule
    \multirow{2}*{\thead{Clip \\ size $\tau$}}& \multirow{2}*{\thead{Clip \\ stride $S$}} & \multirow{2}*{$K_{temp}$} & \multicolumn{3}{c|}{YT-VIS 19~\cite{Yang2019vis} } & \multicolumn{2}{c|}{YT-VIS 21~\cite{Yang2019vis}} & \multicolumn{2}{c}{OVIS~\cite{ovis}} \\
    
    \cmidrule{4-9}
    
    %  & AP & AP$_{75}$ & AP$_{50}$ & AP & AP$_{75}$ & AP$_{50}$ & AP & AP$_{75}$ & AP$_{50}$ \\
%   \midrule
%  0  &$41.2$ &$45.4$  &$63.8$ &$39.5$ &$42.7$  &$61.7$ &$19.7$ &$18.8$  &$39.3$ \\
%   4  &$44.1$ &$48.1$  &$65.1$ &$41.9$ &$46.0$  &$64.8$ &$23.2$ &$21.7$ &$44.4$\\
  
  & & & FPS & AP & $\Delta$ AP & AP & $\Delta$ AP & AP & $\Delta$ AP \\
  \midrule
  6 & 4 & 4 & 18.4 & 44.4 & -- & 43.1 & -- & 23.8 & --\\
  \midrule
  
  3 & 1 & 4 &16.3 & 41.0 & -3.4 &-- & -- &-- & --\\
  9 & 7 & 4 &34.1  & 42.4 & -2.0 &-- & -- &-- & --\\
  12 & 10 & 4 & 36.6 & 41.6 & -2.8 &-- & -- &-- & --\\
%   \midrule
%   6 & 1 & 4 & X & X & X &-- & -- &-- & --\\
%   6 & 2 & 4 & X & X & X &-- & -- &-- & --\\
%   6 & 3 & 4 & X & X & X &-- & -- &-- & --\\
%   6 & 5 & 4 & X & X & X &-- & -- &-- & --\\
    \midrule
  6 & 4 & 0 & 57.7 & 41.2 & -3.2 & 39.5 & -3.6 & 19.7 & -4.1 \\
  6 & 4 & 1 & 40.1 & 41.2 & -3.2 &-- & -- &-- & --\\
  6 & 4 & 2 & 38.8 & 43.4 & -1.0 &-- & -- &-- & --\\
  6 & 4 & 3 & 31.1 & 43.6 & -0.8 &-- & -- &-- & --\\
    
  \bottomrule
 \end{tabular}
 
\end{table*}

\section{Qualitative results}

\label{sec:qual_results_appendix}

In Figure~\ref{fig:qual_results_youtube} and~\ref{fig:qual_results_ovis}, we present additional qualitative results for the YouTube-VIS 2019/2021 and OVIS, respectively.
In Figure~\ref{fig:qual_results_instance_aware_queries}, we demonstrate the instance-aware object queries and their reference point alignment.
To this end, we visualize the attention maps with reference points (cross) at the first, third and last layer for the query from the third frame.
The reference point from that query on other frames aligns with the bounding box positions on these respective frames.
Finally, we present failure cases in Figure~\ref{fig:qual_failure_category} and~\ref{fig:qual_failure_segm}.

\begin{figure}[t]
    \centering
    \includegraphics[width=1\textwidth]{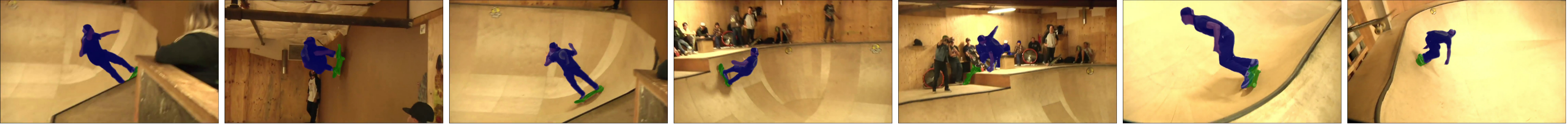}
    \includegraphics[width=1\textwidth]{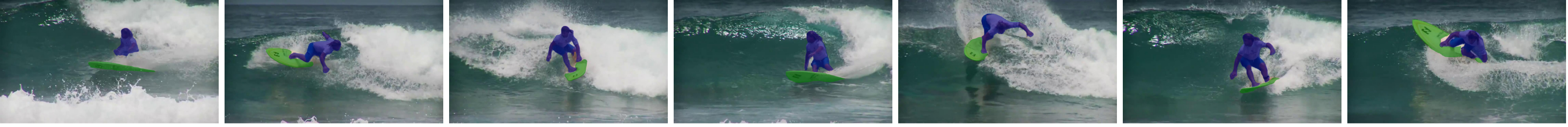}
    \includegraphics[width=1\textwidth]{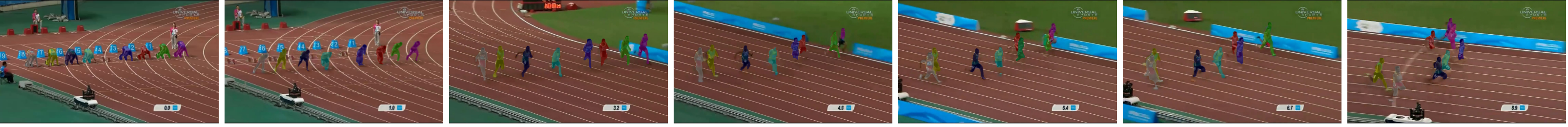}
    \caption{
    Qualitative results on the YouTube-VIS 2019/2021~\cite{Yang2019vis} datasets.
    }
    \label{fig:qual_results_youtube}
\end{figure}
%\vspace{-3pt}
\begin{figure}
    \centering
    \includegraphics[width=1\textwidth]{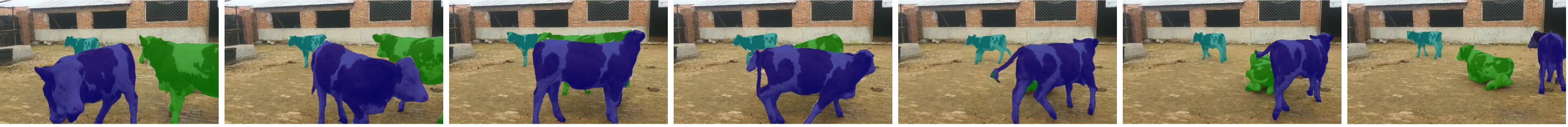}
    \includegraphics[width=1\textwidth]{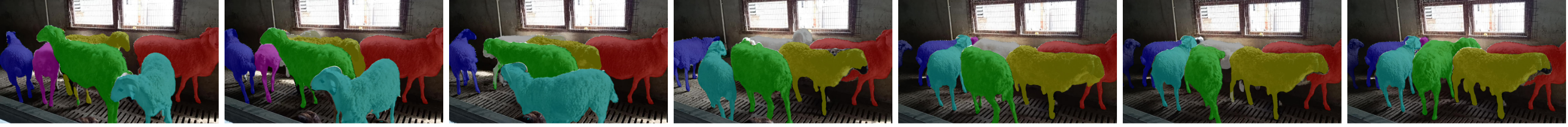}
    \includegraphics[width=1\textwidth]{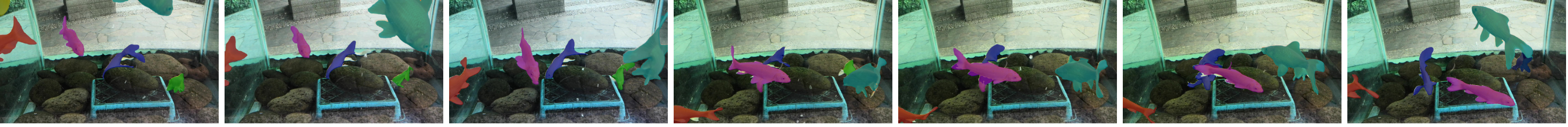}
    \caption{
    Qualitative results on the OVIS~\cite{ovis} dataset.
    }
    \label{fig:qual_results_ovis}
\end{figure}
%\vspace{-3pt}

\begin{figure}
    \centering
    \includegraphics[width=1\textwidth]{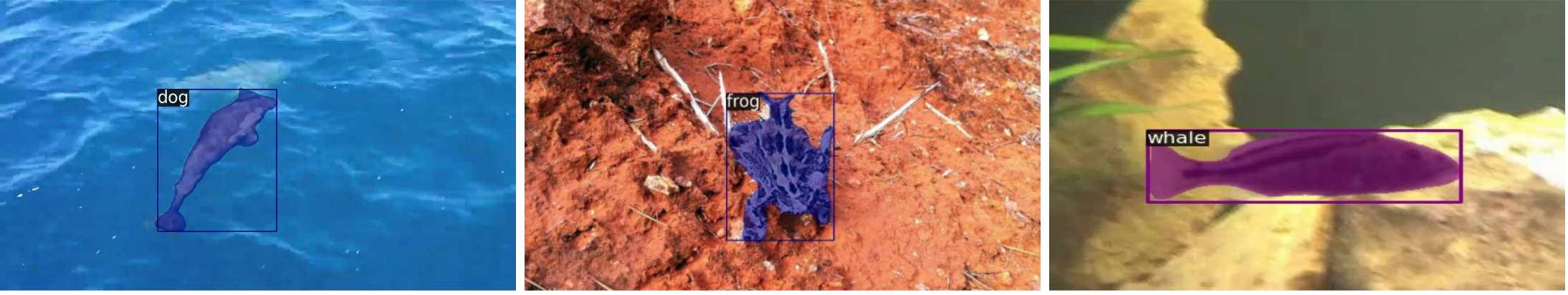}
    \caption{
    Failure predictions due to the category. Most of our errors on YouTube-VIS 2019/2021~\cite{Yang2019vis} datasets are from failed categories, specially the ones that are more under-represented on the training data. 
    }
    \label{fig:qual_failure_category}
\end{figure}
\begin{figure}
    \centering
    \includegraphics[width=1\textwidth]{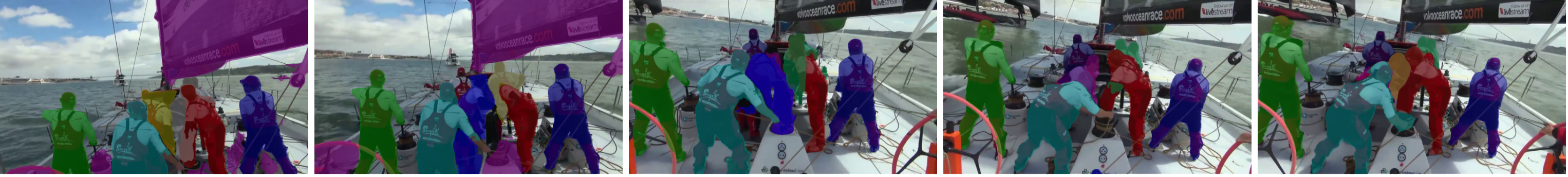}
    \caption{
    Segmentation failure on YouTube-VIS 2021~\cite{Yang2019vis} dataset. We also struggle sometimes to segment overlapping instances from the same category on YouTube-VIS dataset. We do a better job on similar scenarios \ref{fig:qual_results_ovis} on OVIS, and we argue it is because the model sees a lot more of these examples during training in this other dataset. 
    }
    \label{fig:qual_failure_segm}

\end{figure}

\begin{figure}
    \centering
    \includegraphics[width=1\textwidth]{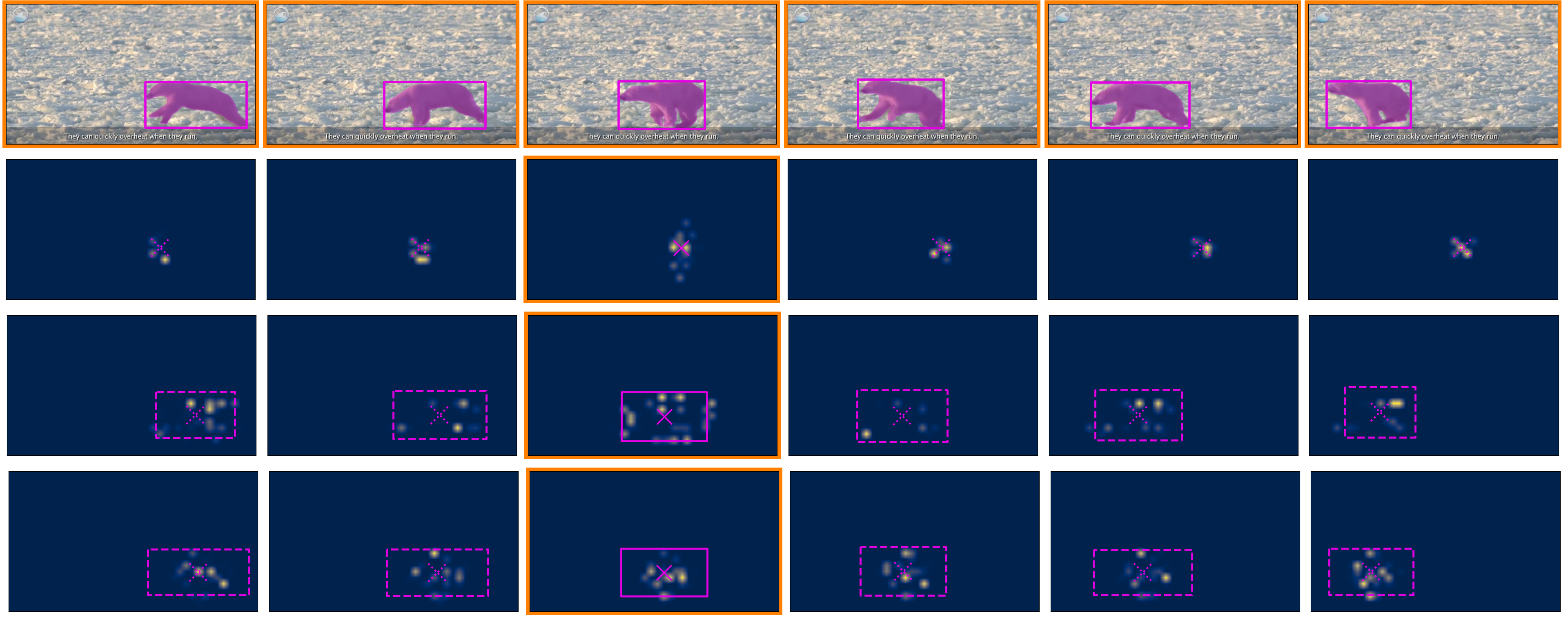}\\
    \vspace{+0.7cm}
    \includegraphics[width=1\textwidth]{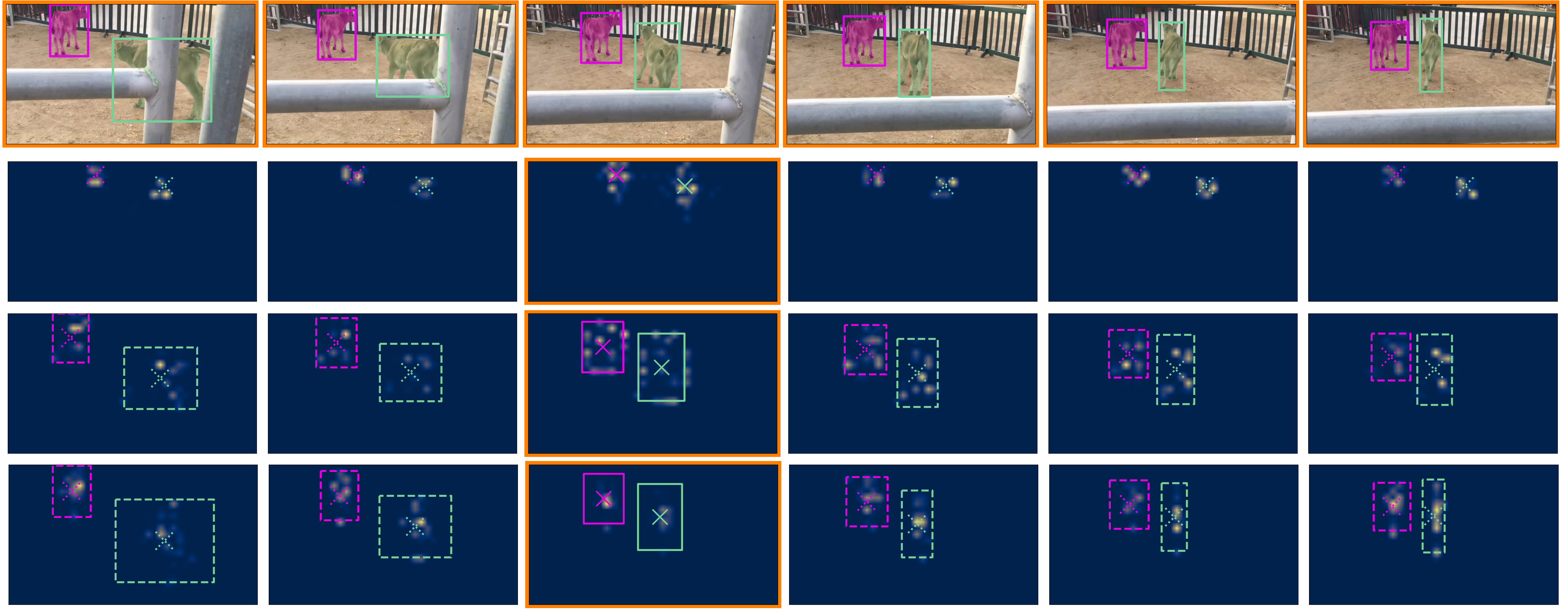}

    \caption{
    Visualization of the instance-aware object queries and their alignment of reference points for temporal frames on example sequences from~\cite{Yang2019vis}.
    To this end, we plot the not yet aligned reference points applied in the first decoder layer (first row) and their subsequent alignment to the predicted bounding box centers after the \nth{2} and \nth{6} layer.
    }
    \label{fig:qual_results_instance_aware_queries}
\end{figure}
%\vspace{-3pt}

% \noindent \textbf{Instance-aware object queries}

% In Figure~\ref{fig:qual_results_instance_aware_queries}

% add figure which demonstrates this with attention maps after first and second decoder level.

% the reference point in the att map from the first decoder layer are not yet aligned.

% in the maps after the second layer u then see the alignment to the bbox pos from the first layer etc.

% \noindent \textbf{Failure cases}

\fi

\clearpage
\bibliographystyle{splncs04}
\bibliography{egbib}

\end{document}